%% file: paper.tex
\documentclass[]{style}
\input{preamble}

\usepackage[utf8]{inputenc} % allow utf-8 input
\usepackage[T1]{fontenc}    % use 8-bit T1 fonts
\usepackage{graphicx}
\usepackage{amsmath}
\usepackage{amssymb}
\usepackage{url}            % simple URL typesetting
\usepackage{booktabs}       % professional-quality tables
\usepackage{amsfonts}       % blacBkboard math symbols
\usepackage{nicefrac}       % compact symbols for 1/2, etc.
\usepackage{xcolor}         % colors
\usepackage{multirow}
\usepackage{comment}
\usepackage{colortbl}
\usepackage{tocloft}
\usepackage{algorithm}
\usepackage{algorithmicx}
\usepackage{algpseudocode}
\usepackage{wrapfig}
\usepackage{tabularx}
\usepackage{textcomp}
\usepackage{stfloats}
\usepackage{verbatim}
\usepackage{titlesec}
\usepackage{adjustbox}
\usepackage{pifont}
\usepackage{tikz}
\usepackage{color}
\usepackage{subcaption}
\usepackage{soul}

\usepackage{makecell}
\newcolumntype{Y}{>{\centering\arraybackslash}X}
\usepackage{multirow}
\usepackage[table]{xcolor} % 提供表格颜色支持
\definecolor{bestcolor}{HTML}{A9D18E} 
\definecolor{sbestcolor}{HTML}{E2EFDA}

\newcolumntype{C}{>{\centering\arraybackslash}m{1.3cm}}
\newcolumntype{Z}{>{\raggedright\arraybackslash\fontsize{6.4pt}{7.2pt}\selectfont}X}

\RequirePackage{xspace}
\makeatletter
\DeclareRobustCommand\onedot{\futurelet\@let@token\@onedot}
\def\@onedot{\ifx\@let@token.\else.\null\fi\xspace}

\makeatother

\definecolor{lightblue}{rgb}{0.66, 0.85, 0.95}
\hypersetup{
    colorlinks=true,
    linkcolor=red,
    citecolor=blue,
}
\definecolor{c2}{HTML}{FBD9BD}
\definecolor{c3}{HTML}{fe793d}
\definecolor{c4}{HTML}{eedeb0}
\definecolor{rouse}{rgb}{0.981,0.961,0.941}
\usepackage[most]{tcolorbox}
\tcbset{colback=blue!5!white, colframe=blue!20!white, boxrule=0pt, arc=2mm, left=1mm, right=1mm, top=1mm, bottom=1mm}

\definecolor{adptorange}{RGB}{248, 205, 172}
\definecolor{cmpblue}{RGB}{189, 215, 238}

\definecolor{our_red}{RGB}{232,157,160}
\definecolor{our_blue}{RGB}{136,206,230}
\definecolor{our_orange}{RGB}{246,200,168}
\definecolor{our_green}{RGB}{178,211,164}

\definecolor{attn_code0}{RGB}{247,215,200}
\definecolor{attn_code1}{RGB}{238,169,139}
\definecolor{mlp_code0}{RGB}{204,201,221}
\definecolor{mlp_code1}{RGB}{102,95,153}

\definecolor{token_blue}{RGB}{84, 120, 140}
\usepackage{pifont}       % \ding{xx}
\usepackage{bbding}       % \Checkmark,\XSolid,... (需要和pifont宏包共同使用)
\usepackage{fontawesome}
\usepackage{xspace}

\usepackage{float}

\newcommand{\ours}{MoRoute\xspace}

\newlength\savewidth

\newcolumntype{x}[1]{>{\centering\arraybackslash}p{#1pt}}
\newcolumntype{y}[1]{>{\raggedright\arraybackslash}p{#1pt}}
\newcolumntype{z}[1]{>{\raggedleft\arraybackslash}p{#1pt}}

\renewcommand{\paragraph}[1]{\vspace{1mm}\noindent\textbf{#1}}

\usepackage{colortbl}
\usepackage{xcolor}
\usepackage{wrapfig}

\renewcommand{\paragraph}[1]{\vspace{1.25mm}\noindent\textbf{#1}}

\usepackage{listings}

\definecolor{codeblue}{rgb}{0.21, 0.49, 0.74}
\definecolor{codekw}{rgb}{0.35, 0.35, 0.75}
\lstdefinestyle{Pytorch}{
    language = Python,
    backgroundcolor = \color{white},
    basicstyle = \fontsize{9pt}{8pt}\selectfont\ttfamily\bfseries,
    columns = fullflexible,
    aboveskip=1pt,
    belowskip=1pt,
    breaklines = true,
    captionpos = b,
    commentstyle = \color{codeblue},
    keywordstyle = \color{codekw},
}

\definecolor{green}{HTML}{009000}
\definecolor{red}{HTML}{ea4335}

\title{MoRoute: Dynamic Routing for In-Context Multimodal Video Generation}
\author[1,2]{Chong Gao}
\author[2]{Jie Ma}
\author[2,3]{Zhan Peng}
\author[2]{Chongxiao Wang}
\author[2]{Haoxue Wu}
\author[2]{Jun Liang}
\author[1,*]{Guanbin Li}
\author[2]{Jing Li}

\affiliation[1]{Sun Yat-sen University}
\affiliation[2]{Orange Team, Moku Lab, HUJING Digital Media \& Entertainment Group}
\affiliation[3]{Huazhong University of Science and Technology}

\contribution[*]{Corresponding author}

\abstract{
Multimodal video generation aims to generate and edit videos conditioned on arbitrary combinations of text, images, and videos within a single model, allowing diverse tasks to share complementary data and generative priors.
Unifying these tasks requires multimodal understanding of diverse conditions, which is typically provided by a pretrained vision-language model (VLM).
A key challenge is how to connect the VLM's hierarchical multimodal representations with a pretrained video diffusion transformer (DiT).
Existing methods either inject features from only the final or a few manually selected VLM layers, or jointly train architecture-matched understanding and generation streams, making it difficult to reuse heterogeneous pretrained backbones.
We introduce \textbf{\ours}, a unified multimodal video generation framework that formulates a frozen VLM and a pretrained video DiT with different architectures as heterogeneous experts connected through dynamic layer routing.
For each input, a lightweight block-wise router enables every DiT block to select the VLM layer most relevant to its generation stage, thereby learning an adaptive correspondence between multimodal understanding and video synthesis.
\ours further incorporates reference images and source videos directly into the DiT token sequence through unified in-context conditioning, preserving fine-grained visual details across diverse generation and editing tasks.
Experiments on IntelligentVBench, OpenVE-Bench, and RefVIE-Bench show that \ours consistently surpasses the best competing method on each benchmark, improving the average score by 0.15, 0.18, and 0.34 on a 1--5 scale, respectively.
}
\project{\url{https://orange-3dv-team.github.io/MoRoute/}}
\date{\today}

\begin{document}
\thispagestyle{firstheader}
\maketitle
\pagestyle{plain}

\input{sec/1_intro}
\input{sec/2_related_work}
\input{sec/3_method}
\input{sec/4_experiment}

\input{sec/5_conclusion}

\clearpage
{
\small
\bibliographystyle{IEEEtran}
\bibliography{main}
}
% \bibliographystyle{assets/plainnat}
% \bibliography{paper}

\clearpage
\beginappendix
\input{sec/X_suppl}

\end{document}

%% file: preamble.tex
\definecolor{lightblue}{rgb}{0.0, 0.6, 1.0}
\definecolor{darkgreen}{rgb}{0.0, 0.6, 0.0}
\definecolor{lightgreen}{rgb}{0.1, 0.8, 0.1}
\definecolor{lightred}{rgb}{0.7, 0.7, 0.7}
\definecolor{lightgreen}{rgb}{0, 0, 0}
\definecolor{lightlightgray}{rgb}{0.8, 0.8, 0.8}

\usepackage{titlesec}
\titlespacing*{\paragraph}
{0pt}{0em}{0.2em}  % {left}{before}{after} spacings

\makeatletter
\renewcommand\paragraph{\@startsection{paragraph}{4}{\z@}%
  {3pt}% 段前间距
  {-0.5em}% 段后间距（负值保证标题与正文在同一行）
  {\normalfont\normalsize\bfseries}} % 格式
\makeatother

\usepackage{amsmath}  % For mathematical symbols
\usepackage{upgreek}
\usepackage{array}    % For better array formatting
\usepackage{multirow}
\usepackage{array}
\usepackage{algorithm}
\usepackage{makecell}
\usepackage{graphicx}
\usepackage{algorithm}
\usepackage{algpseudocode}
\usepackage{colortbl}
\usepackage{enumitem}
\usepackage{wrapfig}
\usepackage{mathrsfs}
\usepackage{pifont}
\usepackage{soul}
\usepackage{amssymb}
\usepackage{tcolorbox}
\usepackage{dirtytalk}
\usepackage{xspace}
\usepackage[utf8]{inputenc}

\usepackage{booktabs}

\usepackage{url}
\definecolor{blue}{rgb}{0.21,0.49,0.74}
\definecolor{red}{rgb}{0.8, 0.2, 0.2}
\definecolor{green}{rgb}{0, 0.5, 0}
\definecolor{yellow}{RGB}{218, 160, 109}
\definecolor{gray}{RGB}{155, 155, 155}

\crefname{section}{Sec.}{Secs.}
\Crefname{section}{Section}{Sections}
\Crefname{table}{Table}{Tables}
\crefname{table}{Tab.}{Tabs.}
\crefname{figure}{Fig.}{Figs.}
\Crefname{figure}{Figure}{Figures}
\crefname{appendix}{App.}{Apps.}
\Crefname{appendix}{Appendix}{Appendices}

%% file: sec/1_intro.tex
\section{Introduction}
\label{sec:intro}

\begin{figure*}[t]
    \centering
    \includegraphics[width=1\textwidth]{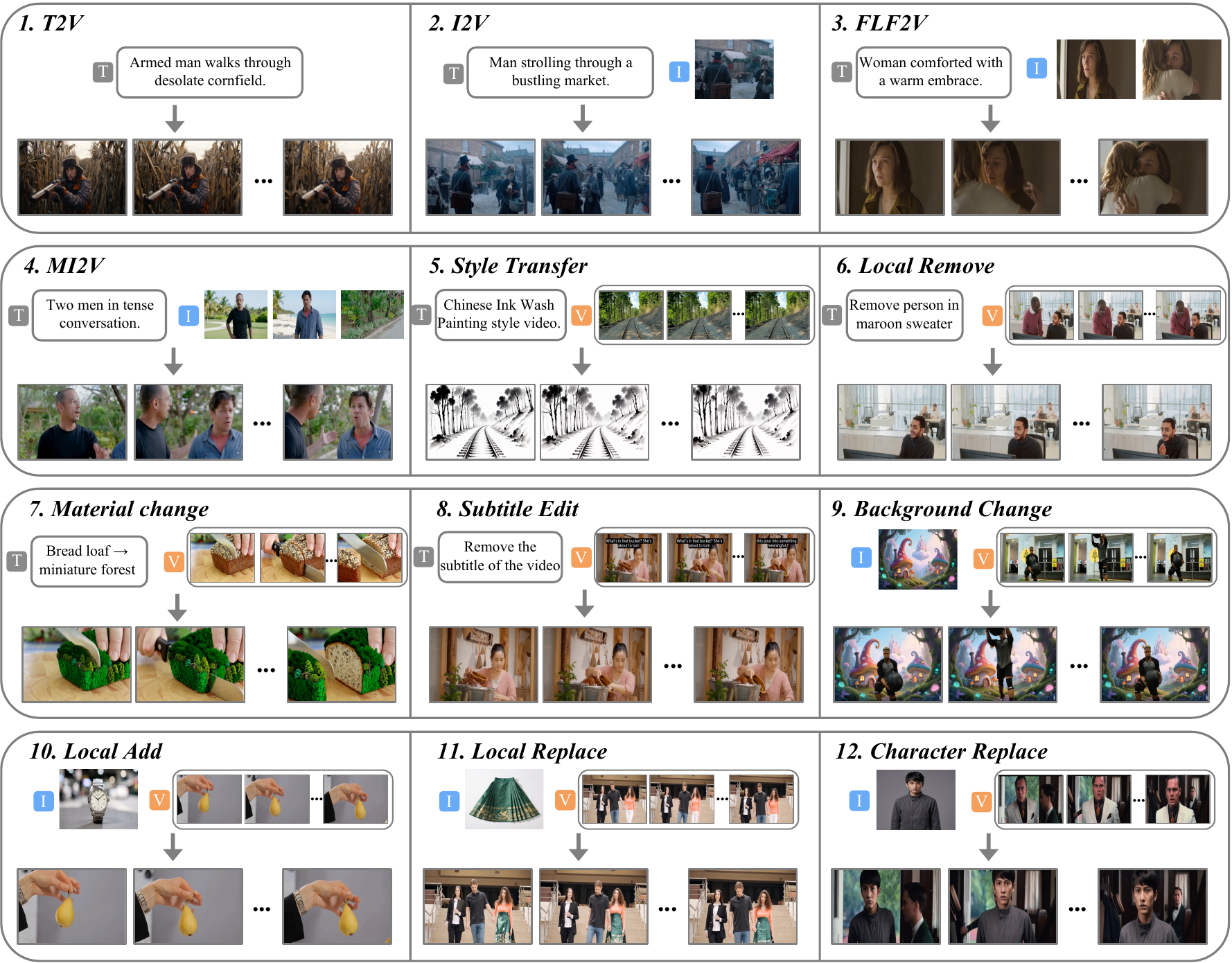}
    \caption{\textbf{\ours supports unified multimodal video generation.}
    The same model handles text-, image-, and video-conditioned tasks, including text-to-video generation, keyframe-to-video generation, reference-guided generation, style transfer, and source-conditioned video editing.
    Each example shows the input conditions and the generated first, middle, and last frames.
    T, I, and V denote text, image, and video conditions, respectively.}
    \label{fig:teaser}
\end{figure*}

Large-scale latent diffusion transformers (DiTs)~\cite{wan2025open,kong2024hunyuanvideo,yang2025cogvideox} have greatly advanced video generation.
Yet practical video creation is inherently multimodal: it increasingly demands control beyond text prompts or first frames, such as reference images that specify subject identity, keyframes that anchor temporal structure, source videos that ground editing, or a mixture of these controls (Figure~\ref{fig:teaser}).
To provide such control, the prevailing practice is to train a dedicated model for each control task, each with its own network architecture and curated dataset.
Although effective for its target task, this one-model-per-task paradigm isolates learning: each model observes only the slice of the real-video distribution covered by its task-specific data~\cite{yang2026context}.
Character replacement video generation is a concrete example. This task lacks real paired data: models are trained on synthetic triples of reference image, source video, and target video, and inherit synthetic artifacts.
Yet jointly training the same model with reference-to-video generation, which has real supervision, yields noticeably cleaner results without ever seeing a real paired replacement sample (Appendix~\ref{supp:motivation}).

Unifying diverse tasks in a single model requires strong multimodal understanding, but most open-source video DiTs are primarily conditioned on text encoders and lack native multimodal understanding.
Existing methods introduce such understanding through two main paradigms.
The first connects a pretrained VLM to a video DiT, reusing the VLM's multimodal representations and the DiT's generation prior.
However, these methods typically use only the final VLM layer or a few hand-picked layers~\cite{lin2026kiwiedit,mou2025instructx,pan2026omniweaving}, leaving much of the VLM's hierarchical representations underused.
The second adopts a Mixture-of-Transformers (MoT) architecture~\cite{deng2025emerging,fu2026lance}, which enables layer-wise interaction between understanding and generation streams.
However, existing MoT designs generally assume matched architecture and jointly train both streams from scratch, precluding the reuse of separately pretrained backbones.

To address these limitations, we introduce \textbf{\ours}, which leverages a \emph{heterogeneous} MoT for multimodal video generation: a frozen VLM expert and a pretrained DiT expert with different architectures, connected by learned layer routing.
The core idea is to let each DiT block adaptively select the VLM layer it needs according to the input and the task.
Specifically, we freeze Qwen3.5-9B~\cite{qwen3.5} as the multimodal condition encoder and equip the Wan2.1-T2V-14B~\cite{wan2025open} DiT with a lightweight per-block router.
To further preserve the dense spatial details of the reference images and source videos,
\ours adopts in-context conditioning: these visual signals are concatenated directly into the DiT input sequence.
Experiments on IntelligentVBench, OpenVE-Bench, and RefVIE-Bench show that \ours achieves the best average score on every evaluated task group, spanning image-conditioned generation, video editing, and reference-guided editing.

\textbf{Our contributions are summarized as follows:}
\begin{itemize}
\item \textbf{Omni-task Video Generation Framework.} We present \textbf{\ours}, a unified framework for text-, image-, and video-conditioned generation, where joint multi-task training allows diverse tasks to share complementary data and generative priors.
\item \textbf{Heterogeneous MoT with Dynamic Routing.} We formulate the frozen VLM and the pretrained DiT as heterogeneous experts in an MoT framework, where a learnable router lets each DiT block dynamically retrieve features from its most relevant VLM layer, reusing both pretrained backbones without architectural constraints.
\item \textbf{Unified In-Context Conditioning.} We design an in-context conditioning architecture that concatenates reference images and source videos directly into the DiT input sequence, ensuring fine-grained visual fidelity while supporting diverse generation and editing tasks.
\end{itemize}

%% file: sec/2_related_work.tex
\section{Related Work}
\label{sec:related_work}

\subsection{Video Diffusion Transformers}

Video generation has moved from U-Net-based diffusion models to large-scale latent DiTs~\cite{wan2025open,kong2024hunyuanvideo,yang2025cogvideox}.
However, most existing systems~\cite{wan2025open,kong2024hunyuanvideo,yang2025cogvideox,hacohen2024ltx,zheng2024open} still rely on frozen text or language encoders such as T5~\cite{raffel2020exploring}, CLIP~\cite{radford2021learning}, or LLaMA-based models.
This text-centric interface works for text-to-video generation but has two limitations for multimodal video generation.
First, these encoders lack the joint multimodal reasoning needed to interpret complex conditions, such as an edit instruction grounded by a reference image and a source video, so visual conditions require separate, task-specific pathways.
Second, the encoder output is shared by all DiT blocks, injected either as cross-attention context or as tokens placed alongside visual tokens; the conditioning thus cannot be specialized to individual blocks.

\subsection{Multimodal Video Generation Frameworks}

Recent methods replace frozen text encoders with VLMs to provide richer multimodal conditioning for video diffusion models.
One line of work extracts a single VLM representation, typically the final-layer hidden states, and feeds it to the DiT through query connectors~\cite{lin2026kiwiedit} or lightweight MLPs~\cite{mou2025instructx}.
Another line designs adapters or condition bridges that aggregate multimodal signals before injection, such as MLLM vision heads~\cite{tan2026omnivideo}, understanding--generation stream connectors~\cite{wei2026univideo}, caption-mediated adapters~\cite{yang2026omnivideo2}, or ViT semantic bottlenecks~\cite{bernini2026latent}.
The most closely related work is OmniWeaving's DeepStacking~\cite{pan2026omniweaving}, which extracts hidden states from multiple VLM layers and injects them into early DiT blocks via additive residuals.
However, its layer selection is manually fixed and independent of the input, so the layer correspondence cannot adapt to block depth or condition type.
Beyond connecting a VLM to a DiT, native unified models such as BAGEL~\cite{deng2025emerging}, Lance~\cite{fu2026lance}, Cosmos~3~\cite{agarwal2026cosmos}, and Omni~\cite{yang2026context} use the LLM backbone itself as the denoiser, coupling homogeneous understanding and generation streams that interact at every layer.
However, this requires both streams to share matched architectures and to be co-trained from scratch, precluding the reuse of separately pretrained VLM and video DiT backbones.
Condition-interface designs such as VACE~\cite{jiang2025vace}, ACE~\cite{han2024ace}, OmniTransfer~\cite{zhang2026omnitransfer}, and OmniShow~\cite{zhou2026omnishow} retain a pretrained DiT and standardize how visual conditions are formatted before injection, but do not involve a VLM for multimodal understanding.

In summary, unified multimodal video generation demands effective use of the VLM's hierarchical representations while preserving both pretrained backbones.
\ours achieves this through block-wise dynamic VLM layer routing: each DiT block uses a lightweight router to select its most relevant VLM layer, forming an input-dependent layer correspondence between two heterogeneous, separately pretrained experts.
A related routing idea appears in Mixture of States (MoS)~\cite{liu2026mixture} for image generation; \ours extends it to video and reuses separately pretrained backbones rather than training from scratch.

%% file: sec/3_method.tex
\section{Method}
\label{sec:method}

\begin{figure*}[t]
    \centering
    \includegraphics[width=1\textwidth]{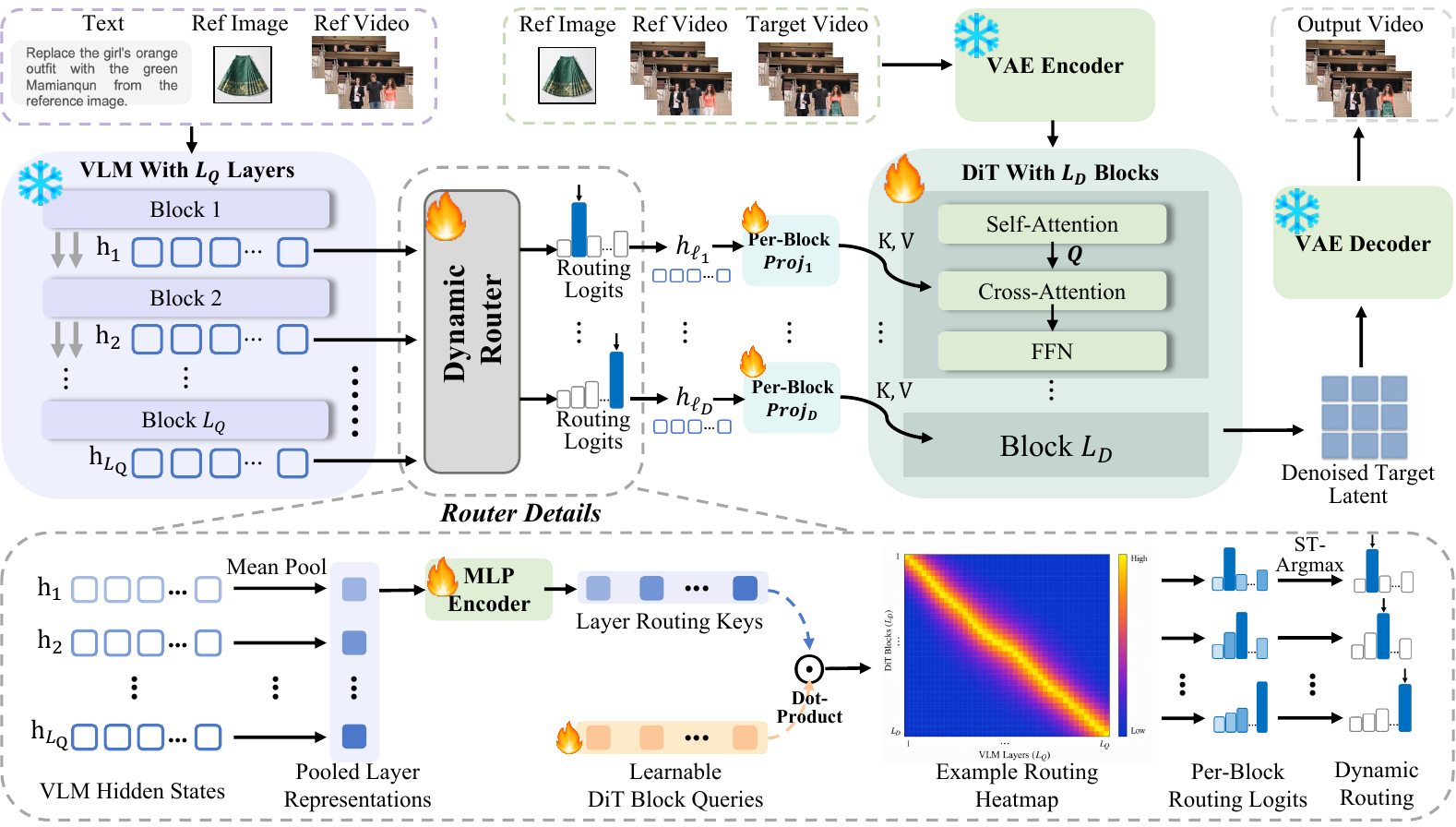}
    \caption{\textbf{Overview of \ours.}
    A frozen VLM encodes multimodal conditions, a video DiT generates the video, and a block-wise dynamic layer router connects them.
    For each DiT block $i$, the router computes dot-product logits between a learnable block query and per-layer VLM keys, and selects VLM layer $\ell_i$ using straight-through argmax. A dedicated per-block projection MLP maps the selected hidden state into the DiT cross-attention context.
    The bottom panel details the router computation, from pooled VLM representations and block queries to the routing heatmap, per-block routing logits, and one-hot selections.
    Snowflake and flame icons denote frozen and trainable modules, respectively.}
    \label{fig:overview}
\end{figure*}

\subsection{Overview}
\label{sec:overview}

We cast \ours as a heterogeneous MoT with two experts of different architectures: a VLM that encodes multimodal conditions and a DiT that generates video.
The VLM is \textbf{Qwen3.5-9B}~\cite{qwen3.5}, a native multimodal vision-language model with $L_Q{=}32$ transformer layers and hidden dimension $d_h{=}4096$.
The DiT is \textbf{Wan2.1-T2V-14B}~\cite{wan2025open}, a latent DiT with $L_D{=}40$ blocks and hidden dimension $d_d{=}5120$.
We first introduce block-wise dynamic layer routing, which connects the two separately pretrained experts by selecting one VLM layer for each DiT block (Sec.~\ref{sec:routing}).
We then present unified in-context conditioning, which injects reference images and source videos directly into the DiT token sequence (Sec.~\ref{sec:multi_condition}).
Finally, we describe the multi-task training corpus of real and synthetic paired videos that supports unified training across generation and editing tasks (Sec.~\ref{sec:training_data}).
Figure~\ref{fig:overview} gives an overview of the framework.

\subsection{Block-wise Dynamic VLM Layer Routing}
\label{sec:routing}

A frozen VLM encodes multimodal inputs into layered representations that range from low-level visual details to high-level semantics.
We introduce a lightweight trainable router that lets each of the $L_D$ DiT blocks select a VLM layer for each input, as illustrated in the bottom panel of Figure~\ref{fig:overview}.

\paragraph{Per-layer representation encoding.}
Given a condition input (text, or text with images/videos), the frozen Qwen3.5 produces hidden states $\{\mathbf{h}_j \in \mathbb{R}^{S \times d_h}\}_{j=1}^{L_Q}$ from its $L_Q$ transformer layers.
We summarize each layer with a compact key $\mathbf{k}_j \in \mathbb{R}^{d_r}$: condition tokens are content-masked mean-pooled, added to a learnable layer-identity embedding, and encoded by a shared MLP; the keys are stacked into $\mathbf{K} \in \mathbb{R}^{L_Q \times d_r}$ (Appendix~\ref{supp:routing}).

\paragraph{Per-block queries and logit computation.}
We associate each DiT block $i$ with a learnable query vector $\mathbf{q}_i \in \mathbb{R}^{d_r}$.
Routing logits are computed by a normalized dot product between DiT queries and VLM layer keys:
\begin{equation}
  L_{i,j}
  =
    \frac{\mathbf{q}_i}{\|\mathbf{q}_i\|_2}
    \cdot
    \frac{\mathbf{k}_{j}}{\|\mathbf{k}_{j}\|_2},
  \qquad
  \mathbf{L} \in \mathbb{R}^{L_D \times L_Q},
  \label{eq:logits}
\end{equation}
where the $L_2$ normalization bounds the learned logits and stabilizes routing (the exact logit scaling is given in Appendix~\ref{supp:routing}).

\paragraph{Gaussian position prior.}
The dot-product logits alone do not encode the depth order of VLM and DiT blocks, which can make early training unstable or collapse the routing.
We therefore add a fixed near-diagonal Gaussian bias $b_{i,j}$ that maps each DiT block to its proportional position among the $L_Q$ VLM layers, encouraging shallow blocks to attend to shallow layers early in training.
The final logits combine the learned logits with this prior:
\begin{equation}
  L^{\mathrm{final}}_{i,j} = L_{i,j} + \alpha \, b_{i,j},
  \qquad
  \mathbf{L}^{\mathrm{final}} \in \mathbb{R}^{L_D \times L_Q},
  \label{eq:logits_final}
\end{equation}
where $\alpha$ is a bias scale annealed by a cosine schedule, letting the router gradually rely on the learned logits.

\paragraph{Discrete selection via straight-through argmax.}
For DiT block $i$, the selected VLM layer is $\ell_i = \arg\max_j L^{\mathrm{final}}_{i,j}$, with soft routing distribution $\mathbf{p}_i = \mathrm{softmax}(\mathbf{L}^{\mathrm{final}}_i)$.
Since direct argmax is non-differentiable, training uses a straight-through estimator that forwards the one-hot selection of $\ell_i$ while backpropagating through $\mathbf{p}_i$ (Appendix~\ref{supp:routing}).
At inference, we use $\ell_i$ to index the selected VLM hidden state directly.

\paragraph{Per-block condition projection.}
For each DiT block $i$, the selected hidden state $\mathbf{h}_{\ell_i}$ is projected from $\mathbb{R}^{S \times d_h}$ to $\mathbb{R}^{S \times d_d}$ by a dedicated MLP $\mathrm{Proj}_i$ with layer normalization and GELU activation, yielding the block-specific context $\mathbf{c}_i$.
The $L_D$ projections do \emph{not} share weights, so each DiT block can learn its own adaptation from the selected VLM representation (projection MLP structure in Appendix~\ref{supp:routing}; parameter and compute overhead of the interface in Appendix~\ref{supp:training}).

\paragraph{Hierarchical injection into the DiT.}
The resulting context vectors $\{\mathbf{c}_{i}\}_{i=1}^{L_D}$ are fed to the corresponding DiT blocks as cross-attention context, while the remaining block components (timestep-conditioned AdaLN, self-attention, and feed-forward) follow the original DiT.

\paragraph{Training Objectives.}
The router is trained jointly with the DiT under the flow-matching objective $\mathcal{L}_{\mathrm{FM}}$.
We add three auxiliary terms to shape the routing: a \emph{diversity} loss $\mathcal{L}_{\mathrm{div}}$ that keeps the aggregate VLM layer usage from collapsing onto a few layers, a \emph{confidence} loss $\mathcal{L}_{\mathrm{conf}}$ that sharpens each block's top-1 selection, and a light \emph{monotonicity} regularizer $\mathcal{L}_{\mathrm{mono}}$ that encourages a shallow-to-deep routing tendency (full definitions in Appendix~\ref{supp:objectives}), yielding the total objective:
\begin{equation}
  \mathcal{L} = \mathcal{L}_{\mathrm{FM}}
    + \lambda_{\mathrm{div}}  \, \mathcal{L}_{\mathrm{div}}
    + \lambda_{\mathrm{conf}} \, \mathcal{L}_{\mathrm{conf}}
    + \lambda_{\mathrm{mono}} \, \mathcal{L}_{\mathrm{mono}}.
  \label{eq:total_loss}
\end{equation}

\subsection{In-Context Multi-Condition Video Diffusion}
\label{sec:multi_condition}

The routing interface (Sec.~\ref{sec:routing}) determines \emph{how} the VLM conditions the DiT, but not \emph{what} signals are provided as input for each task.
Tasks differ primarily in their input conditions: text prompts alone, added keyframes or images, reference subjects, or complete source videos.
We unify these disparate inputs as \emph{in-context video generation}: all available reference images and source videos are directly concatenated into the DiT input sequence.
These reference signals carry dense spatial and textural details that might be compromised by cross-attention alone, but are effectively preserved through in-context concatenation.

\subsubsection{Multi-condition Token Concatenation with Slotted Temporal RoPE}
\label{sec:token_concat}

Following common practice~\cite{jiang2025vace,wan2025open}, we patchify all visual inputs and concatenate them into one token sequence.
Given a target latent $\mathbf{z}_{\mathrm{tgt}}$,
up to $K_v$ conditioning videos $\{\mathbf{z}^{(j)}_{\mathrm{cv}}\}$,
and up to $K_i$ conditioning images $\{\mathbf{z}^{(k)}_{\mathrm{ci}}\}$:
\begin{equation}
    \mathbf{x} = [\,\underbrace{\mathrm{Patch}(\mathbf{z}_{\mathrm{tgt}})}_{\text{target}}\;;\;\underbrace{\mathrm{Patch}(\mathbf{z}^{(1)}_{\mathrm{cv}})\;;\;\cdots}_{\text{cond videos}}\;;\;\underbrace{\mathrm{Patch}(\mathbf{z}^{(1)}_{\mathrm{ci}})\;;\;\cdots}_{\text{cond images}}\,].
\end{equation}

The key challenge is position encoding, because input segments may have different resolutions and temporal lengths.
We use \emph{slotted temporal RoPE}: each segment occupies a non-overlapping range on the temporal axis of the 3D rotary position embedding, while all segments share the same spatial RoPE basis (illustrated in Appendix~\ref{supp:incontext}).
The target occupies $t \in [0, T_{\mathrm{tgt}})$, conditioning videos are placed at offsets of 100 (e.g., $t \in [100, 100{+}T_{c_1})$), and each conditioning image occupies a single-frame slot starting from $t = 300$, giving each segment a distinct positional identity.

\subsubsection{Sparse Attention}
\label{sec:sparse_attn}

Token concatenation lets the target and conditions exchange information through self-attention, but full attention over the concatenated sequence is costly as the number of conditions grows.
Since conditioning inputs are clean, we adopt a sparse pattern in which target tokens attend to all tokens while each conditioning segment attends only to itself.
This preserves spatial coherence within each clean segment while reducing attention cost (Appendix~\ref{supp:incontext}).

\subsubsection{Dual Timestep Modulation}
\label{sec:dual_timestep}

The concatenated sequence contains two noise levels: target latents are corrupted at timestep $t$, while conditioning inputs remain clean.
Standard DiT blocks derive one AdaLN modulation from $t$ and apply it to all tokens, mixing these two signals.
We address this with \emph{dual timestep modulation}: each DiT block computes AdaLN parameters twice, once from $t$ for target tokens and once from $t_{\mathrm{cond}} = 0$ for conditioning tokens.
Since condition tokens always use $t_{\mathrm{cond}} = 0$, their timestep embeddings and modulation parameters are independent of the denoising step and can be precomputed and cached outside the denoising loop at inference, speeding up generation.

\subsection{Training Data}
\label{sec:training_data}

We use large-scale open-source text-to-image datasets~\cite{schuhmann2022laion} to establish initial semantic alignment between the VLM's representations and the DiT's generative space.
For multimodal video generation, we construct a training corpus from two complementary sources: \textbf{real video data} and \textbf{synthetic paired video data}.

\paragraph{Real video data.}
We collect large-scale open-source video corpora, including Vchitect-T2V-Dataverse~\cite{fan2025vchitect}, and generate detailed captions using VLMs.
For MI2V tasks, we employ Qwen3.5-9B to extract key concepts from the video and use SAM3~\cite{carion2025sam} for corresponding segmentation.
Existing methods typically crop a portion of a frame or regenerate reference images with image editing models, causing copy-paste artifacts or reference--target discrepancies.
We instead propose a \textit{cross-frame} strategy: rather than cropping or regenerating within the same frame, we take a segment from a longer video as the training target and draw reference images from different frames.
The reference thus shares the same underlying concepts with enough visual variation to prevent overfitting to exact pixel-level matches.

\paragraph{Synthetic paired video data.}
Video editing requires paired before-and-after examples, which are scarce in the real world and thus primarily obtained through synthetic generation.
We leverage public editing corpora~\cite{lin2026kiwiedit,bai2026scaling,fu2026EffectErase} for operations including object removal, effect erasing, style transfer, and local modification, and further use Unreal Engine 5 (UE5) to generate synthetic pairs for character replacement, animation, and camera motion transfer.

%% file: sec/4_experiment.tex
\section{Experiments}

\subsection{Implementation Details}
\label{sec:training_details}

\paragraph{Training and inference.}
\label{sec:training_pipeline}
Model configurations follow Sec.~\ref{sec:overview}, with video latents encoded by a frozen Wan2.1 VAE.
We train the model with a three-stage progressive curriculum: (i) text-to-image alignment of the condition interface on LAION-2B~\cite{schuhmann2022laion} with the DiT frozen, (ii) text-to-video adaptation of the unfrozen DiT, and (iii) multi-condition training on a mixture of T2V, TI2V, TV2V, and TIV2V tasks using the architecture in Sec.~\ref{sec:multi_condition}.
All stages use flow matching with bf16 mixed precision on 16 H20 GPUs, at $480 \times 832$ resolution with 81 frames for video.
At inference, we generate videos using 50 denoising steps and a classifier-free guidance scale of 5.0.
Routing hyperparameters and per-stage data and optimization settings are detailed in Appendix~\ref{supp:training}.

\subsection{Experimental Settings}

\paragraph{Evaluation benchmarks.}
We evaluate \ours on three benchmarks covering multimodal video generation and editing:
\textbf{IntelligentVBench}~\cite{pan2026omniweaving} covers Compositional Multi-Image-to-Video (MI2V), Implicit Image-to-Video (I2V), Interpolative Dual-Image-to-Video (DI2V), and Text-Image-Video-to-Video (TIV2V) with local replacement, background change, and object addition.
\textbf{OpenVE-Bench}~\cite{lin2026kiwiedit} focuses on Text-Video-to-Video (TV2V) editing. It includes background change, camera edit, creative edit, global style transfer, local add/change/remove, and subtitle edit.
\textbf{RefVIE-Bench}~\cite{lin2026kiwiedit} targets reference-guided video editing.

\paragraph{Compared methods.}
We compare with five methods representing different condition injection paradigms (Sec.~\ref{sec:related_work}): OmniWeaving~\cite{pan2026omniweaving}, Bernini~\cite{bernini2026latent}, Kiwi-Edit~\cite{lin2026kiwiedit}, Omni-Video~2~\cite{yang2026omnivideo2}, and VACE~\cite{jiang2025vace}.
Since not all methods support every condition format, we report comparisons on each method's supported tasks; in particular, Kiwi-Edit and Omni-Video~2 do not support I2V generation and are omitted from those subtasks.

\paragraph{Evaluation protocol.}
Following the benchmark protocols~\cite{pan2026omniweaving,lin2026kiwiedit}, we use Gemini-2.5-Pro as the automatic evaluator with the official prompts and 1--5 scoring criteria, running all baselines from their official codebases under this shared protocol.
Each benchmark reports its own task-specific dimensions (defined in Table~\ref{tab:main_results} and Appendix~\ref{supp:full_tables}).

\subsection{Main Results}
\label{sec:main_results}

We organize results by task family rather than by benchmark.
Table~\ref{tab:main_results} summarizes the results on the three benchmarks (full per-subtask and per-editing-type breakdowns in Appendix~\ref{supp:full_tables}), and Figure~\ref{fig:qual_main} shows qualitative results.

\begin{table*}[tb]
\centering
\small
\setlength{\tabcolsep}{4pt}
\begin{tabular}{l|cccc|cccc|ccc}
\toprule
 & \multicolumn{4}{c|}{\textbf{IntelligentVBench}} & \multicolumn{4}{c|}{\textbf{OpenVE-Bench}} & \multicolumn{3}{c}{\textbf{RefVIE-Bench}} \\
\textbf{Method} & IF$\uparrow$ & CP$\uparrow$ & VQ$\uparrow$ & AVG$\uparrow$ & IC$\uparrow$ & CD$\uparrow$ & VQ$\uparrow$ & AVG$\uparrow$ & Subj.$\uparrow$ & BG$\uparrow$ & AVG$\uparrow$ \\
\midrule
VACE             & 3.16 & 3.07 & 3.23 & 3.15 & 1.52 & 1.52 & 1.52 & 1.52 & 1.00 & 1.40 & 1.20 \\
Bernini          & 3.79 & 3.75 & 3.46 & 3.67 & \underline{3.93} & \underline{3.49} & \underline{3.54} & \underline{3.65} & \underline{3.94} & \underline{3.60} & \underline{3.77} \\
Kiwi-Edit$^\dagger$   & \cellcolor{gray!25}3.89 & \cellcolor{gray!25}3.78 & \cellcolor{gray!25}3.67 & \cellcolor{gray!25}3.78 & 3.28 & 2.82 & 2.99 & 3.03 & 3.00 & 2.53 & 2.77 \\
Omni-Video~2$^\dagger$ & \cellcolor{gray!25}3.00 & \cellcolor{gray!25}2.33 & \cellcolor{gray!25}3.22 & \cellcolor{gray!25}2.85 & 3.28 & 3.02 & 3.12 & 3.14 & 3.17 & 2.27 & 2.72 \\
OmniWeaving      & \underline{3.92} & \underline{3.81} & \underline{3.76} & \underline{3.83} & 3.16 & 2.67 & 2.65 & 2.82 & 3.11 & 3.33 & 3.22 \\
\textbf{\ours}   & \textbf{4.08} & \textbf{4.02} & \textbf{3.84} & \textbf{3.98} & \textbf{4.07} & \textbf{3.66} & \textbf{3.76} & \textbf{3.83} & \textbf{4.43} & \textbf{3.78} & \textbf{4.11} \\
\bottomrule
\end{tabular}
\caption{Main quantitative comparison across the three benchmarks.
IntelligentVBench reports IF (Instruction Following), CP (Condition Preserving), and VQ (Visual Quality), averaged over its four subtasks;
OpenVE-Bench reports IC (Instruction Compliance), CD (Consistency and Detail Fidelity), and VQ (Visual Quality and Stability), averaged over its eight editing types;
RefVIE-Bench reports the subject- (Subj.) and background-editing (BG) category averages, with AVG their mean.
$^\dagger$: the IntelligentVBench columns cover the TIV2V subset only, as I2V-style generation is unsupported; these entries (gray cells) are excluded from the IntelligentVBench ranking.
Full results in Appendix~\ref{supp:full_tables}.
Best results are \textbf{bold}; second best are \underline{underlined}.}
\label{tab:main_results}
\end{table*}

% ============================================================
\paragraph{TI2V / I2V-style generation.}
This family comes from IntelligentVBench and includes three subtasks:
\textbf{Compositional MI2V} integrates multiple reference subjects into a specified background and follows the described actions;
\textbf{Implicit I2V} performs image-to-video generation with implicit text instructions that may introduce elements not visible in the input image;
and \textbf{Interpolative DI2V} generates a temporally coherent transition conditioned on both the first and last frames.
\ours achieves the highest AVG on all three subtasks (Appendix~\ref{supp:full_tables}) and the best IntelligentVBench average overall (Table~\ref{tab:main_results}).

% ============================================================
\paragraph{TV2V editing.}
This family comes from OpenVE-Bench and covers several editing operations on source videos:
\textbf{global style} transfer across the entire video, \textbf{background change} while keeping foreground subjects intact, \textbf{local add/change/remove} of specific objects, \textbf{camera edit} of motion or viewpoint, \textbf{creative edit} with stylized transformations, and \textbf{subtitle edit} of text overlays.
\ours obtains the best overall AVG on OpenVE-Bench (Table~\ref{tab:main_results}) and leads most editing types in the per-type breakdown (Appendix~\ref{supp:full_tables}).

% ============================================================
\paragraph{TIV2V generation/editing.}
This family comes from RefVIE-Bench and the TIV2V subset of IntelligentVBench.
It is the most challenging setting because the model must use a text instruction, a reference image, and a source video at the same time:
\textbf{local replacement} of an object in the source video with the subject from the reference image, \textbf{background change} to the scene depicted in the reference image, and \textbf{object addition} inserting the reference subject into a designated location within the video.
\ours achieves the best AVG on RefVIE-Subject, RefVIE-Background, and IntelligentVBench TIV2V (Appendix~\ref{supp:full_tables}), yielding the best RefVIE-Bench average in Table~\ref{tab:main_results}.
On RefVIE-Background, Bernini has a higher visual-harmony sub-score (Appendix~\ref{supp:full_tables}), but \ours obtains stronger reference fidelity and matting quality, leading to the best category average.

\begin{figure*}[t]
    \centering
    \begin{subfigure}[b]{0.86\linewidth}
        \centering
        \includegraphics[width=\linewidth]{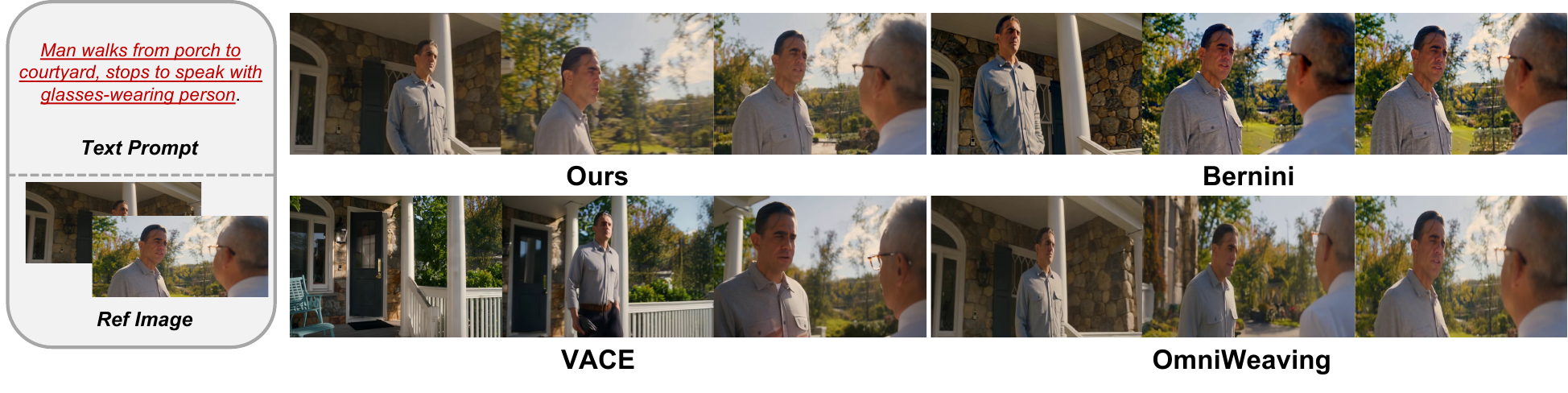}
    \end{subfigure}

    \vspace{6pt}

    \begin{subfigure}[b]{0.86\linewidth}
        \centering
        \includegraphics[width=\linewidth]{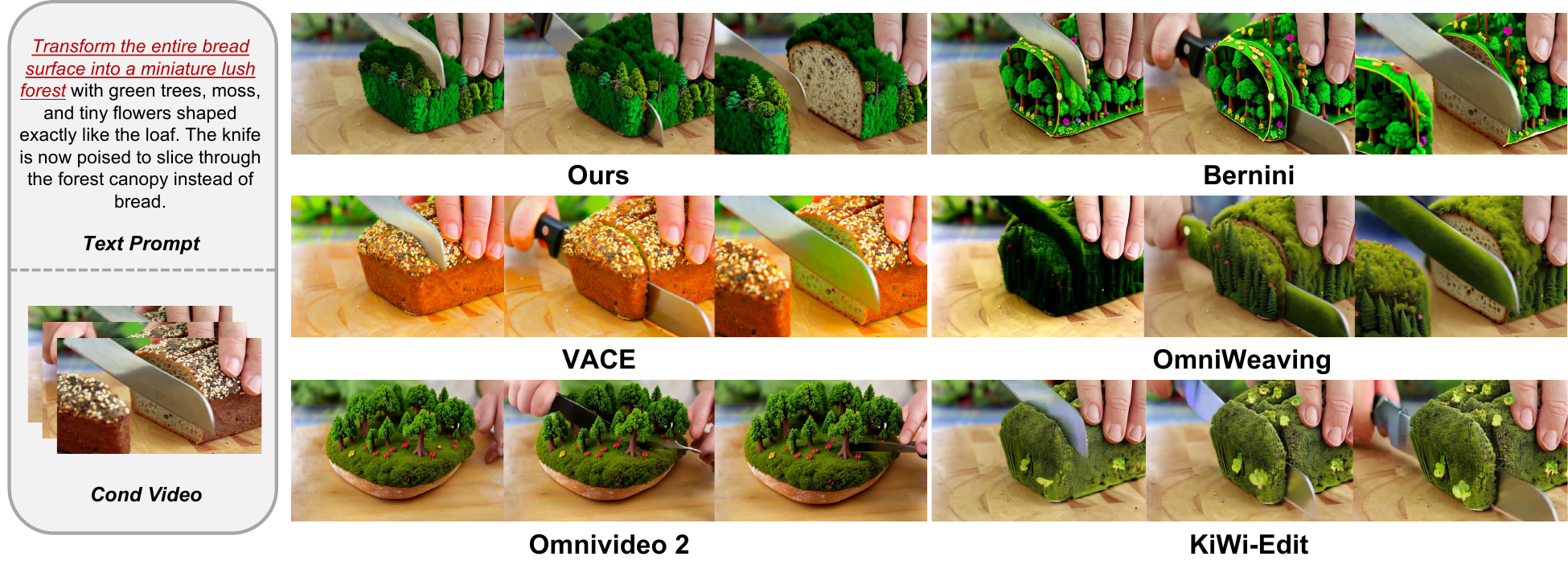}
    \end{subfigure}

    \vspace{6pt}

    \begin{subfigure}[b]{0.86\linewidth}
        \centering
        \includegraphics[width=\linewidth]{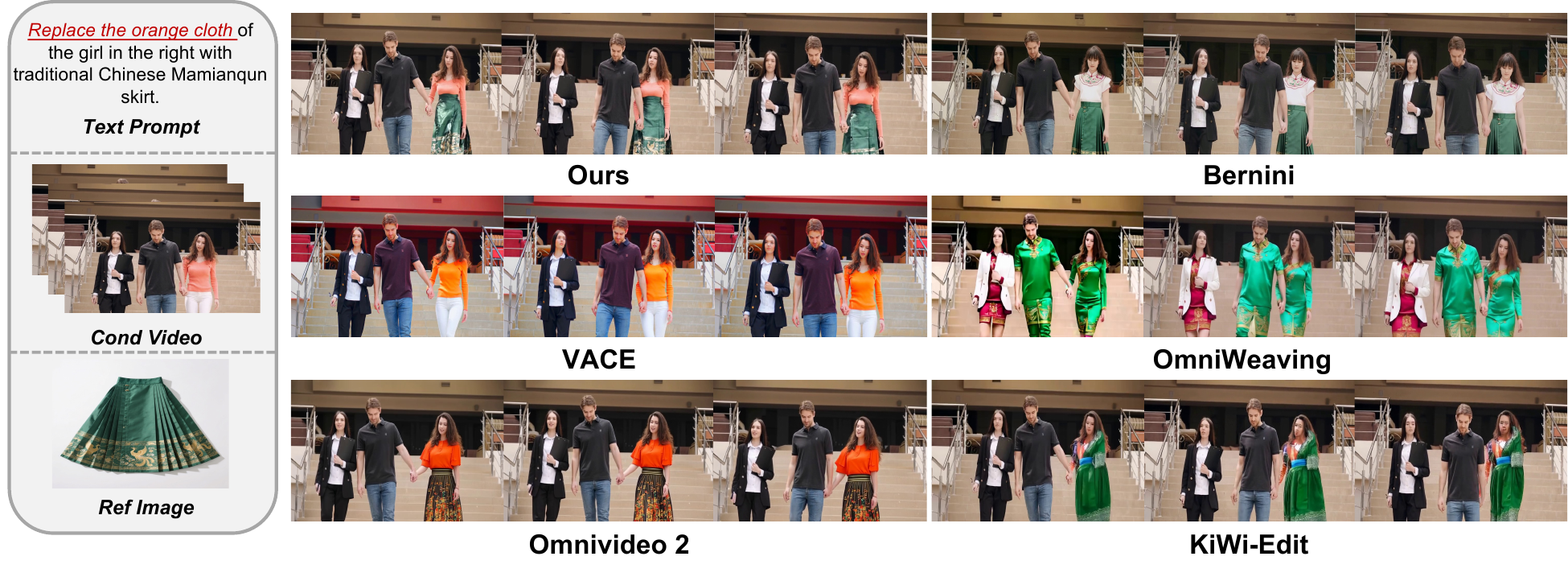}
    \end{subfigure}
    \caption{\textbf{Qualitative comparison across task families.}
    One representative case per family, from top to bottom: TI2V / I2V-style generation, TV2V editing, and TIV2V generation/editing.
    Each case compares \ours with baselines using the first, middle, and last frames.
    Additional cases for each family are provided in Appendix~\ref{supp:gallery}.}
    \label{fig:qual_main}
\end{figure*}

\subsection{Ablation Study}
\label{sec:ablation}

\begin{figure*}[t]
    \centering
    \includegraphics[width=0.86\textwidth]{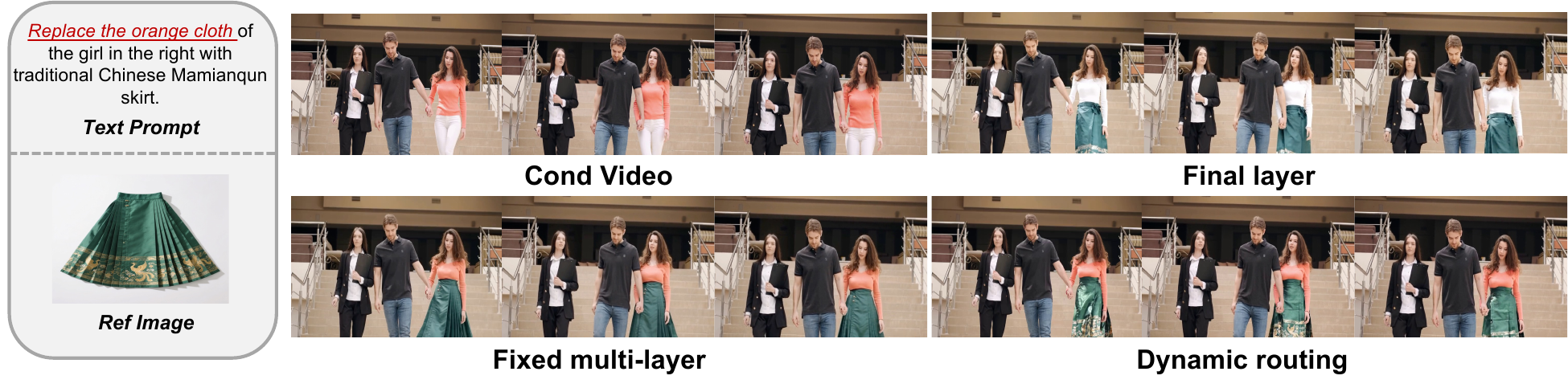}
    \caption{\textbf{Qualitative comparison of routing strategies.} 
    The case shows the prompt and conditions (left), the source video (Cond Video), and the outputs of the three routing strategies.
    Dynamic routing follows the instruction faithfully while the fixed strategies either miss the intended edit or lose source details.}
    \label{fig:ablation_routing}
\end{figure*}

To validate the effectiveness of block-wise dynamic routing, we compare three layer-selection strategies under the same architecture and training recipe:
\textbf{Final layer}, where all DiT blocks share the last VLM layer;
\textbf{Fixed multi-layer}, where blocks follow a uniform diagonal mapping from shallow to deep VLM layers; and
\textbf{Dynamic routing}, our full method.
All variants are trained with the full Stage~1 and shortened Stage~2 and Stage~3 schedules (0.5 epoch each; scores thus not comparable to Table~\ref{tab:main_results}), and evaluated on OpenVE-Bench and the TIV2V subset of IntelligentVBench.
As shown in Table~\ref{tab:ablation_routing}, Fixed multi-layer outperforms Final layer, confirming that intermediate VLM layers carry complementary information beyond the final layer.
Dynamic routing further surpasses Fixed multi-layer on both task groups (+0.12 and +0.15 AVG), demonstrating the benefit of input-conditioned layer selection.
Figure~\ref{fig:ablation_routing} illustrates this qualitatively: dynamic routing preserves fine-grained visual details and follows instructions more faithfully than both fixed strategies (additional qualitative results in Appendix~\ref{supp:gallery}; a task- and instance-level analysis of the learned routing in Appendix~\ref{supp:routing_analysis}).

\begin{table}[tb]
\centering
\small
\setlength{\tabcolsep}{1.5pt}
\begin{tabular}{l|cccc|cccc}
\toprule
 & \multicolumn{4}{c|}{\textbf{OpenVE-Bench}} & \multicolumn{4}{c}{\textbf{IVB TIV2V}} \\
\textbf{Strategy} & IC$\uparrow$ & CD$\uparrow$ & VQ$\uparrow$ & AVG$\uparrow$ & IF$\uparrow$ & CP$\uparrow$ & VQ$\uparrow$ & AVG$\uparrow$ \\
\midrule
Final layer        & 3.42 & 3.50 & 3.55 & 3.49 & 3.82 & 3.80 & 3.86 & 3.83 \\
Fixed multi-layer  & 3.58 & 3.64 & 3.66 & 3.63 & 4.02 & 3.98 & 4.00 & 4.00 \\
\textbf{Dynamic routing} & \textbf{3.70} & \textbf{3.76} & \textbf{3.79} & \textbf{3.75} & \textbf{4.20} & \textbf{4.16} & \textbf{4.10} & \textbf{4.15} \\
\bottomrule
\end{tabular}
\caption{Ablation on VLM layer routing strategies.
IVB TIV2V denotes the TIV2V subset of IntelligentVBench; dimension abbreviations follow Table~\ref{tab:main_results}.
Best results are \textbf{bold}.}
\label{tab:ablation_routing}
\end{table}

%% file: sec/5_conclusion.tex
\section{Conclusion}
\label{sec:conclusion}

We presented \ours, a unified framework for multimodal video generation and editing.
The key idea is to connect a frozen VLM and a pretrained video DiT through block-wise dynamic layer routing: instead of relying on the final VLM layer or a fixed set of layers, each DiT block selects the VLM layer most relevant to the current input, more effectively leveraging the VLM's hierarchical representations.
We also introduced an in-context multi-condition video diffusion architecture that supports text, image, and video conditions in one model.
Slotted temporal RoPE, sparse attention, and dual timestep modulation help the DiT handle clean conditioning tokens and noisy target tokens together, and a progressive three-stage training pipeline stabilizes the heterogeneous interface.
Experiments on IntelligentVBench, OpenVE-Bench, and RefVIE-Bench show that \ours achieves the highest average score on every evaluated task group while leading the majority of individual scoring dimensions.
These results demonstrate that adaptive layer routing and in-context multi-condition design together enable a practical unified system for multimodal video generation and editing.

\paragraph{Limitations and future work.}
\textbf{Broader modalities:} \ours currently handles only text, image, and video conditions; incorporating audio and 3D signals into the same routing interface is a promising direction toward general any-to-any generation.
\textbf{Model scaling:} our study is currently limited to a single pair of backbones, leaving the behavior of dynamic routing with larger models unexplored; scaling the two experts and the training data may further improve overall capability and cross-task generalization.

%% file: sec/X_suppl.tex
% ===========================================================================
% Appendix (adapted from the AAAI supplementary material; cross-references
% now point to the main text within the same document).
% ===========================================================================

% ===========================================================================
\section{Extended Motivation}
\label{supp:motivation}

This section expands on the motivating observation summarized in
Sec.~\ref{sec:intro}: multi-task training with real reference-to-video data
improves a data-scarce task.

Reference-guided character replacement requires the model to insert a subject
specified by a reference image into a source video while preserving the
identity, motion, and scene of the source.
This task lacks real-world paired data: collecting a source video, a reference
image, and the corresponding edited video with a different character identity
is difficult at scale.
As a result, models are typically trained on synthetic triples of reference
image, source video, and target video, and their outputs tend to exhibit
synthetic artifacts inherited from the rendering pipeline.

\begin{figure*}[b]
    \centering
    \includegraphics[width=0.9\textwidth]{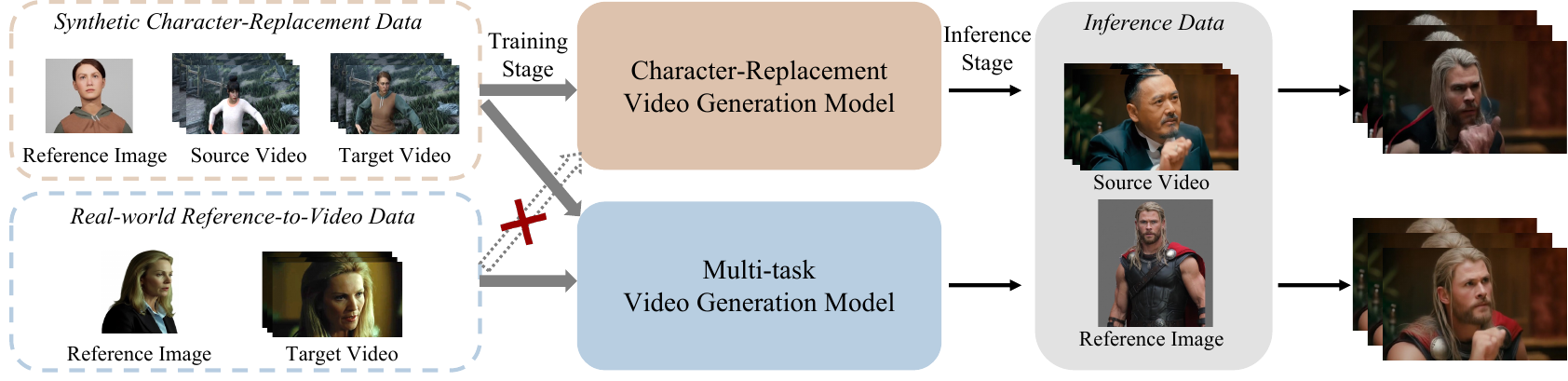}
    \caption{\textbf{Multi-task training with real reference-to-video data improves a data-scarce task.}
    We compare two training settings for character replacement under the same inference input.
    A model trained only on synthetic character-replacement pairs remains limited by synthetic supervision.
    In contrast, multi-task training with real-world reference-to-video data produces a cleaner character-replacement result, even without real-world paired samples for this task.}
    \label{fig:insight}
\end{figure*}

Figure~\ref{fig:insight} contrasts two training settings under the same
inference input.
A model trained only on synthetic character-replacement pairs remains limited
by synthetic supervision and produces visibly synthetic results.
When the same model is jointly trained with a related task that has real
supervision---reference-to-video generation---the character-replacement result
becomes noticeably cleaner, even though no real-world paired sample of
character replacement is ever seen.
This shows that unifying multiple tasks in a single model lets a data-scarce
task borrow real-world priors from data-rich tasks, which is the central
motivation for the unified, multi-task design of \ours.

% ===========================================================================
\section{Routing Interface Details}
\label{supp:routing}

This section provides the encoding, position-prior, and projection details of
the block-wise dynamic VLM layer routing described in Sec.~\ref{sec:routing}.

\subsection{Per-layer Representation Encoding}

Before routing, we represent each VLM layer with a compact key that DiT blocks
can compare against.
Given a condition input (text, or text with images/videos), Qwen3.5 produces
hidden states $\{\mathbf{h}_j \in \mathbb{R}^{S \times d_h}\}_{j=1}^{L_Q}$ from
its $L_Q{=}32$ transformer layers, where $S$ is the sequence length.
We skip the embedding layer output (index 0) and retain only the transformer
layer outputs.

\paragraph{Mean pooling.}
Each VLM layer outputs a token sequence, while the router needs one vector per
layer.
We therefore mean-pool the token features into a layer-level representation.
We first drop the system prompt and padding tokens so that only condition
tokens contribute:
\begin{equation}
  \bar{\mathbf{h}}_j = \frac{\sum_s M_s \cdot \mathbf{h}_{j,s}}{\sum_s M_s},
  \label{eq:mean_pool}
\end{equation}
where $M_s=1$ for condition tokens and $M_s=0$ for the dropped system prompt and padding tokens.

\paragraph{Layer key encoding.}
The pooled vectors summarize the content seen at each VLM depth, but they do not explicitly encode which layer they come from.
We add a learnable layer embedding $\mathbf{e}_j \in \mathbb{R}^{d_h}$ to encode layer identity.
The sum is then passed through a shared MLP to produce a compact routing key:
\begin{equation}
  \mathbf{k}_j =
  \mathrm{MLP}_{\mathrm{key}}\bigl(\bar{\mathbf{h}}_j + \mathbf{e}_j\bigr),
  \qquad
  \mathbf{k}_j \in \mathbb{R}^{d_r},
  \label{eq:layer_key}
\end{equation}
where the same key encoder is shared by all VLM layers.
It uses a GELU hidden layer
($d_h \!\to\! 2d_r \!\to\! d_r$).
The resulting keys are stacked into $\mathbf{K} \in \mathbb{R}^{L_Q \times d_r}$.

\subsection{Gaussian Position Prior}
\label{supp:gaussian_prior}

The dot-product logits between DiT block queries and VLM layer keys do not
encode the depth order of VLM and DiT blocks.
Without a prior, early training can produce unstable or collapsed routing.
We therefore add a fixed diagonal Gaussian bias $\mathbf{B} \in \mathbb{R}^{L_D \times L_Q}$ to the logits:
\begin{equation}
  b_{ij} = -\frac{(j - c_i)^2}{2\sigma^2}, \qquad
  c_i = 1 + (i - 1) \cdot \frac{L_Q - 1}{L_D - 1},
  \label{eq:bias}
\end{equation}
where $i \in \{1,\ldots,L_D\}$ and $j \in \{1,\ldots,L_Q\}$.
The center $c_i$ maps DiT block $i$ to its proportional position among the $L_Q$ VLM layers.
This prior encourages a near-diagonal pattern early in training: shallow DiT blocks attend to shallow VLM layers, while deeper blocks attend to deeper layers.
The prior enters the final logits scaled by a coefficient $\alpha$ (Eq.~\ref{eq:logits_final}), which is annealed with a cosine schedule during early training (Sec.~\ref{supp:routing_setup}); this progressively weakens the bias, so the router gradually relies more on the learned routing logits, which can override the prior as training progresses.

\paragraph{Logit scaling.}
In implementation, the $L_2$-normalized dot-product logits are multiplied by a fixed temperature $\tau$ before the prior is added, bounding the learned term to $[-\tau, \tau]$.
With $\tau{=}40$ (Sec.~\ref{supp:routing_setup}), the learned term dominates the annealed prior at convergence while keeping the routing distribution sharp enough for confident top-1 selection.

\subsection{Straight-Through Layer Selection}

For DiT block $i$, the selected VLM layer is $\ell_i = \arg\max_j L^{\mathrm{final}}_{i,j}$, with soft routing distribution $\mathbf{p}_i = \mathrm{softmax}(\mathbf{L}^{\mathrm{final}}_i)$.
Direct argmax is non-differentiable, so during training we use a straight-through (ST) argmax estimator:
\begin{equation}
  \mathbf{w}_{i} =
    \mathrm{onehot}(\ell_i)
    - \mathrm{sg}(\mathbf{p}_i)
    + \mathbf{p}_i,
  \qquad
  \mathbf{w}_{i} \in \mathbb{R}^{L_Q},
  \label{eq:st_argmax}
\end{equation}
where $\mathrm{sg}(\cdot)$ denotes stop-gradient.
The forward pass uses the one-hot selection of layer $\ell_i$, while the backward pass follows the softmax surrogate $\mathbf{p}_i$.

\subsection{Per-block Condition Projection}

For each DiT block $i$, the selected VLM hidden state is projected from the VLM
dimension to the DiT dimension by a dedicated MLP $\mathrm{Proj}_i$
(Sec.~\ref{sec:routing}).
Each $\mathrm{Proj}_i$ is a two-layer MLP with layer normalization, GELU
activation, and dropout.
Its structure is LayerNorm $\to$ Linear $\to$ GELU $\to$ Linear $\to$ Dropout
$\to$ LayerNorm, with dimensions $d_h \to d_d \to d_d$.
The $L_D$ projections do \emph{not} share weights, so each DiT block can learn
its own adaptation from the selected VLM representation.

% ===========================================================================
\section{Training Objectives}
\label{supp:objectives}

The router is trained jointly with the DiT under the flow-matching objective
$\mathcal{L}_{\mathrm{FM}}$.
As summarized in Sec.~\ref{sec:routing}, we add three auxiliary terms---a
diversity loss, a confidence loss, and a light monotonicity regularizer---to
keep routing diverse, sharp, and softly ordered.
Their definitions follow, where $p_{i,j}$ denotes the $j$-th element of the
soft routing distribution $\mathbf{p}_i = \mathrm{softmax}(\mathbf{L}^{\mathrm{final}}_i)$
for DiT block $i$.

\paragraph{Diversity loss.}
We add an entropy term over the aggregate VLM layer usage distribution.
It keeps the aggregate layer usage from collapsing onto a few VLM layers:
\begin{equation}
  \mathcal{L}_{\mathrm{div}} =
  1 - \frac{H(\mathbf{u})}{\log L_Q},
  \qquad
  u_{j} = \frac{\sum_{i} p_{i,j}}{\sum_{j'}\sum_{i} p_{i,j'}},
  \label{eq:div_loss}
\end{equation}
where $\mathbf{u}$ is the normalized VLM layer usage, and $H(\cdot)$ denotes Shannon entropy.
The normalization by $\log L_Q$ bounds the loss in $[0,1]$; zero means uniform VLM layer usage.

\paragraph{Confidence loss.}
To make each routing decision sharp, we maximize the top-1 softmax probability:
\begin{equation}
  \mathcal{L}_{\mathrm{conf}} =
  -\frac{1}{L_D}\sum_{i=1}^{L_D}
    \log\!\bigl(\max_j \, p_{i,j}\bigr).
  \label{eq:conf_loss}
\end{equation}
This loss reaches zero when every DiT block routes with probability one to a single VLM layer, and increases as the distribution becomes more uniform.

\paragraph{Monotonicity regularizer.}
Finally, we add a light monotonicity regularizer that gently encourages a shallow-to-deep routing tendency:
\begin{equation}
  \mathcal{L}_{\mathrm{mono}} =
  \frac{1}{L_D{-}1}
  \sum_{i=1}^{L_D-1}
    \mathrm{ReLU}\!\bigl(m - (\mu_{i+1} - \mu_{i})\bigr),
  \label{eq:mono_loss}
\end{equation}
where $\mu_i = \sum_j p_{i,j} \cdot j$ is the expected VLM layer index under $\mathbf{p}_i$, and the margin $m \geq 0$ sets the minimum expected increment between adjacent blocks.
Since $(L_D{-}1)\,m = 39$ exceeds the maximum index span $L_Q{-}1 = 31$ under our setting ($m{=}1.0$; Sec.~\ref{supp:routing_setup}), this term cannot reach zero for all block pairs; it acts as a soft ordering pressure that trades off against the other objectives rather than a hard constraint.
These three terms are combined with the flow-matching loss to form the total
objective in Eq.~\ref{eq:total_loss}.

% ===========================================================================
\section{In-Context Conditioning Details}
\label{supp:incontext}

This section details the slotted temporal RoPE and the sparse attention pattern used for in-context multi-condition video diffusion.

\begin{figure}[t]
    \centering
    \begin{subfigure}[b]{0.55\linewidth}
        \centering
        \includegraphics[width=\linewidth]{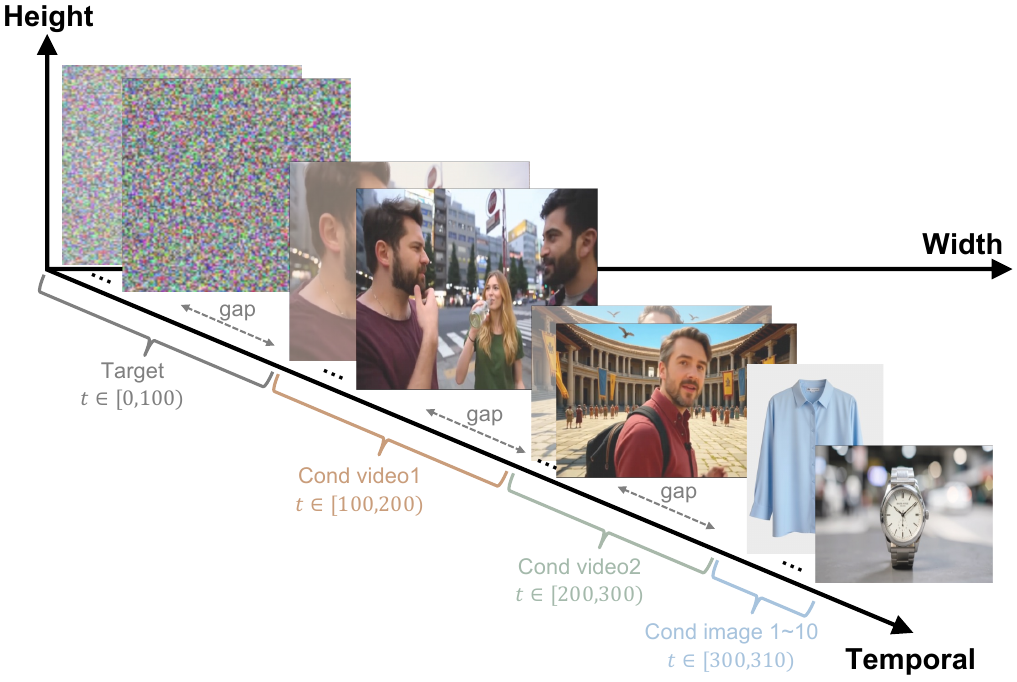}
        \caption{Slotted 3D RoPE.}
        \label{fig:rope}
    \end{subfigure}
    \hfill
    \begin{subfigure}[b]{0.38\linewidth}
        \centering
        \includegraphics[width=\linewidth]{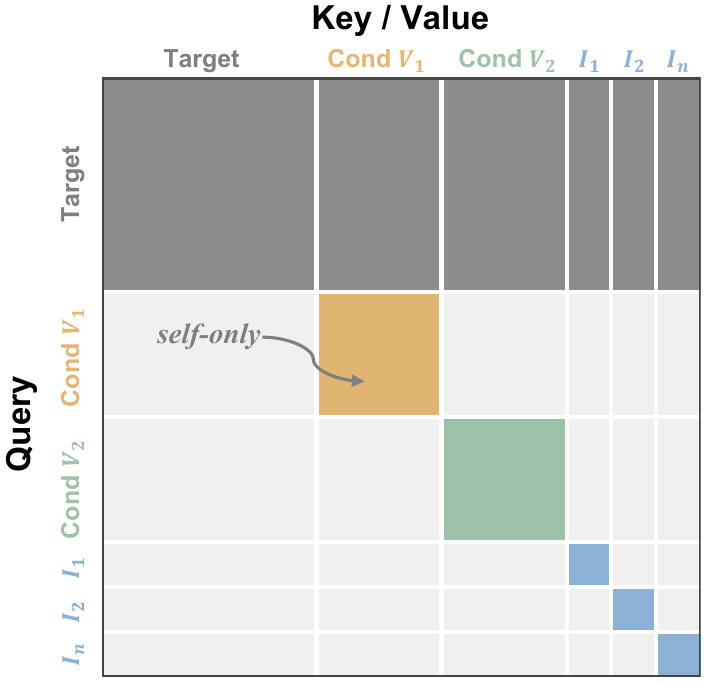}
        \caption{Sparse attention mask.}
        \label{fig:sparse_attn}
    \end{subfigure}
    \caption{In-context conditioning details.
    (a) Slotted 3D RoPE: each segment uses a separate temporal slot while sharing the spatial $(h, w)$ RoPE basis; slots show reserved ranges, and actual occupancy is at most 21 latent frames per video segment.
    (b) Sparse attention mask: target tokens attend to all tokens, while each conditioning segment attends to itself.}
    \label{fig:incontext_details}
\end{figure}

\paragraph{Slotted temporal RoPE.}
Each input segment occupies a non-overlapping range on the temporal axis of the 3D rotary position embedding, while all segments share the same spatial RoPE basis (Figure~\ref{fig:rope}): the target occupies $t \in [0, T_{\mathrm{tgt}})$, conditioning videos are placed at offsets of 100, and each conditioning image occupies a single-frame slot starting from $t = 300$.

\paragraph{Slot capacity and non-overlap.}
The fixed offsets cannot cause overlap because the maximum video length is bounded.
All videos are trained and evaluated with at most 81 frames, which the causal VAE compresses to 21 latent frames ($4\times$ temporal downsampling), so every video segment spans at most 21 temporal positions, far below the slot spacing of 100.
Concretely, the target occupies $t \in [0, 21)$, the $j$-th conditioning video occupies $t \in [100j, 100j{+}21)$, and conditioning images occupy consecutive single-frame slots in the reserved range $t \in [300, 310)$.
This instantiation thus supports up to two conditioning videos and up to ten conditioning images.
The offsets are hyperparameters and can be enlarged proportionally for longer videos.

\paragraph{Sparse attention.}
Token concatenation lets the target and conditions exchange information through self-attention.
However, full attention becomes costly when the sequence length $N = N_{\mathrm{tgt}} + \sum_j N_{c_j} + \sum_k N_{i_k}$ grows, especially with multiple videos.
Conditioning inputs are clean signals, so they do not need to attend to the noisy target or to other conditions.
We therefore adopt a sparse attention pattern with two asymmetric rules (Figure~\ref{fig:sparse_attn}):
\begin{itemize}
    \item \textbf{Target $\to$ All.} Target tokens attend to the entire concatenated sequence, including themselves and all conditioning segments.
    This gives denoising access to all conditions.
    \item \textbf{Condition $\to$ Self only.} Each conditioning segment attends only to its own tokens.
    Since conditions are clean, self-attention within each segment is enough to preserve spatial coherence.
\end{itemize}
This pattern reduces attention complexity from $\mathcal{O}(N^2)$ to $\mathcal{O}\!\left(N_{\mathrm{tgt}} \cdot N + \sum\nolimits_j N_{c_j}^2 + \sum\nolimits_k N_{i_k}^2\right)$.

% ===========================================================================
\section{Training Data and Implementation Details}
\label{supp:training}

\subsection{Multi-task Training Data}

Training \ours requires data for image-level alignment, video generation, and multi-condition video editing.
We use large-scale open-source text-to-image (T2I) datasets to establish initial semantic alignment between the VLM's representations and the DiT's generative space.
For multimodal video generation, we construct a training corpus from two complementary sources: \textbf{real video data} and \textbf{synthetic paired video data}.
We normalize each training example into a target video, a text condition, a possibly empty set of condition images, and a possibly empty set of condition videos.
The generation and editing tasks are instantiated by changing only the available visual conditions.

\paragraph{Real video data.}
We collect large-scale open-source video corpora, including Vchitect-T2V-Dataverse~\cite{fan2025vchitect}, and generate detailed captions using VLMs.
The resulting text--video pairs are used directly for text-to-video (T2V) training.
We derive image-conditioned examples from the same corpus: either the first or last frame forms an image-to-video (I2V) condition, while both endpoint frames are used for first-last-frame-to-video (FLF2V); explicit textual role descriptions indicate whether each image serves as the first or last frame.
For multi-image-to-video (MI2V) tasks, we employ Qwen3.5-9B to extract key concepts from the video and use SAM3~\cite{carion2025sam} for corresponding segmentation.
Existing methods typically crop a portion of a frame or use image editing models to regenerate reference images, leading to copy-paste artifacts or significant discrepancies between the reference and target video.
To address this, we propose a \textit{cross-frame} strategy: instead of cropping or regenerating within the same frame, we select a segment from a longer video as the training target and choose reference images from different video frames.
This ensures that the reference shares the same underlying concepts while maintaining sufficient visual variation, preventing overfitting to exact pixel-level matches.

\paragraph{Synthetic paired video data.}
Video editing requires paired before-and-after examples, which are scarce in the real world and thus primarily obtained through synthetic generation.
We leverage public editing corpora---OpenVE~\cite{lin2026kiwiedit}, Ditto~\cite{bai2026scaling}, and EffectErase~\cite{fu2026EffectErase}---for operations including object removal, effect erasing, style transfer, and local modification, and further use Unreal Engine 5 (UE5) to generate synthetic pairs for character replacement, animation, and camera motion transfer.
The OpenVE editing pairs used for training are drawn from a split disjoint from the OpenVE-Bench test set.
More broadly, no evaluation sample from IntelligentVBench, OpenVE-Bench, or RefVIE-Bench appears in any training split, so all reported results are free of train--test contamination.
These combined sources provide controlled, task-specific examples that are difficult to obtain from natural video collections.

\subsection{Routing Setup}
\label{supp:routing_setup}

The router performs per-input layer selection with a straight-through argmax.
Layer keys are built with content-masked mean pooling followed by a shared two-layer MLP key encoder, the router dimension is $d_r=256$, and the logit temperature is $\tau{=}40$.
The Gaussian prior uses $\sigma^2{=}100.0$, and its scale $\alpha$ decays from 6.25 to 1.0 over the first 10,000 steps of Stage~1, following a cosine schedule on the effective variance (from 16.0 to 100.0); the final scale is kept for the later stages and at inference.
We set the monotonicity loss weight to 1.0 with margin 1.0, the diversity loss weight to 1.0, and the confidence loss weight to 0.3.

\subsection{Interface Overhead}
\label{supp:overhead}

The newly added condition interface consists of the router and the $L_D$ per-block projections; the DiT's cross-attention layers belong to the pretrained backbone and are counted in its 14B parameters.
The router itself is lightweight: the shared key encoder, layer-identity embeddings, and block queries total 2.37M parameters.
The per-block projections dominate the interface size: each $\mathrm{Proj}_i$ maps $d_h{=}4096$ to $d_d{=}5120$, giving 47.2M parameters per block and 1.89B in total ($\approx$13.5\% of the DiT backbone).
Table~\ref{tab:supp_overhead} reports the breakdown.

The runtime overhead is small relative to the DiT.
The router runs once per sample, as its inputs depend only on the condition rather than the denoising step, and each projection is applied only to the condition tokens (a few hundred), whereas DiT self-attention operates on tens of thousands of video tokens; the interface therefore contributes a negligible fraction of per-step FLOPs.
In terms of memory, the interface adds roughly 3.8\,GB of weights in bf16, while its activation overhead is small because the projected condition sequences are orders of magnitude shorter than the video token sequence.

\begin{table}[tb]
\centering
\small
\begin{tabular}{lr}
\toprule
\textbf{Component} & \textbf{Params} \\
\midrule
Router (key encoder, layer embeddings, queries) & 2.37\,M \\
Per-block projections ($L_D{=}40$) & 1.89\,B \\
\midrule
Total condition interface & 1.89\,B \\
DiT backbone & 14\,B \\
\bottomrule
\end{tabular}
\caption{Parameter breakdown of the condition interface.}
\label{tab:supp_overhead}
\end{table}

\subsection{Progressive Training Pipeline}

We divide the learnable parameters into two groups: the newly added \emph{condition interface} (router and per-block projections) together with the DiT's cross-attention layers, and the rest of the DiT backbone.
These modules have different initialization and optimization dynamics, so we train them progressively: first align the condition interface, then adapt the backbone, and finally train the full multi-task model.
All stages are trained on 16 H20 GPUs with 96GB memory per GPU.
All experiments run on Linux with Python~3.10, PyTorch~2.10 with CUDA~12.8, HuggingFace Transformers~4.57, and the Accelerate library, using bf16 mixed precision throughout.

\textbf{Stage~1 (Text-to-Image).}
We first train the condition interface on a 9,900,000-sample subset of LAION-2B~\cite{schuhmann2022laion}.
Images are resized to $512 \times 512$, and each sample is treated as a single-frame input.
Only the router, per-block projections, and cross-attention modules are trainable, while the DiT backbone remains frozen.
We train for 2 epochs with Adam, learning rate $1\times10^{-4}$, bf16 mixed precision, per-device batch size 48, and gradient accumulation 1.
This stage aligns the VLM-to-DiT interface without the added difficulty of temporal modeling.

\textbf{Stage~2 (Text-to-Video).}
We switch to T2V data and unfreeze the full DiT for training.
This stage uses 600,000 T2V samples drawn from Vchitect-T2V-Dataverse~\cite{fan2025vchitect} and internally curated videos, all recaptioned with a unified pipeline.
Videos are trained at $480 \times 832$ resolution with 81 frames.
We train for 2 epochs with learning rate $1\times10^{-5}$ and per-device batch size 1.
With the interface initialized from Stage~1, the DiT backbone learns temporal modeling while adapting to the per-block context.

\textbf{Stage~3 (Multi-condition).}
We train on a mixture of text-to-video (T2V), text-image-to-video (TI2V), text-video-to-video (TV2V), and text-image-video-to-video (TIV2V) tasks using the multi-condition architecture.
The data mixture contains 600K T2V samples, 600K TI2V samples (300K FLF2V/I2V and 300K MI2V), 600K TV2V samples, and 1M TIV2V samples.
The TV2V and TIV2V data are drawn from OpenVE~\cite{lin2026kiwiedit}, Ditto~\cite{bai2026scaling}, EffectErase~\cite{fu2026EffectErase}, and UE5 character-replacement sources.
The sampling ratio is 0.2 for T2V, 0.2 for TI2V, 0.3 for TV2V, and 0.3 for TIV2V.
We use $480 \times 832$ resolution, 81 frames, learning rate $5\times10^{-6}$, per-device batch size 1, and train for 2 epochs.
This stage teaches the model to handle different conditioning inputs in a single framework and supports cross-task synergy.

This curriculum moves from static alignment to temporal generation and then to multi-task unification.

\paragraph{Inference settings.}
At inference time, we generate videos at $480 \times 832$ resolution with 81 frames and 24 fps.
We use 50 denoising steps, a classifier-free guidance (CFG) scale of 5.0, and a fixed random seed for all benchmark runs.

% ===========================================================================
\section{Full Quantitative Results}
\label{supp:full_tables}

This appendix reports the full quantitative results underlying the
benchmark-level averages in Table~\ref{tab:main_results}.
Table~\ref{tab:supp_intelligent_vbench} reports the per-dimension results on
the three IntelligentVBench generation subtasks;
Table~\ref{tab:supp_openve} reports the per-editing-type results on
OpenVE-Bench;
Table~\ref{tab:supp_refvie_tiv2v} reports the per-dimension results on
RefVIE-Bench and the TIV2V subset of IntelligentVBench.
Kiwi-Edit and Omni-Video~2 do not support I2V-style generation and are
therefore absent from Table~\ref{tab:supp_intelligent_vbench}.

\paragraph{Evaluation protocol.}
All scores are produced by Gemini-2.5-Pro following each benchmark's official evaluation prompts and scoring criteria; we do not modify the prompts or the score scales.
For each generated video, the judge receives the task instruction, the conditioning inputs, and frames uniformly sampled at 6\,fps, and returns per-dimension scores on the benchmark's five-point scale.
Each video is scored three times and the scores are averaged, and all methods are evaluated with identical prompts, frame sampling, and judge settings.

\paragraph{Baseline setup.}
All baselines are run from their official code releases with the publicly released checkpoints and default inference settings: VACE (Wan2.1-VACE-14B), Bernini (Bernini-Diffusers, 7B{+}14B), Kiwi-Edit (kiwi-edit-5b-instruct-reference-diffusers), Omni-Video~2 (OmniVideo2-A14B), and OmniWeaving (HY-OmniWeaving).
Each method receives the same task instruction and conditioning inputs and generates videos at its native resolution and length; no post-processing is applied before scoring.

On OpenVE-Bench, \ours achieves the best score on seven of the eight editing
types (tying with Bernini on Local Remove).
The main exception is Camera Edit, where Bernini scores highest (4.80
vs.\ 3.82), showing that camera-specific control remains challenging for the
unified interface and is a direction for targeted improvement.

\begin{table*}[tb]
\centering
\small
\setlength{\tabcolsep}{3pt}
\begin{tabular}{l|cccc|cccc|cccc}
\toprule
 & \multicolumn{4}{c|}{\textbf{Compositional MI2V}} & \multicolumn{4}{c|}{\textbf{Implicit I2V}} & \multicolumn{4}{c}{\textbf{Interpolative DI2V}} \\
\textbf{Method} & IF$\uparrow$ & CP$\uparrow$ & VQ$\uparrow$ & AVG$\uparrow$ & IF$\uparrow$ & CP$\uparrow$ & VQ$\uparrow$ & AVG$\uparrow$ & IF$\uparrow$ & CP$\uparrow$ & VQ$\uparrow$ & AVG$\uparrow$ \\
\midrule
VACE             & 3.43 & 3.56 & 3.71 & 3.57 & 3.60  & 3.60 & 3.00 & 3.40 & 3.71 & 3.00 & 3.43 & 3.38 \\
Bernini          & 3.67 & 3.76 & \underline{3.87} & 3.77 & 3.67 & 3.67 & 2.67 & 3.34 & 3.71 & 3.57 & 3.29 & 3.52 \\
OmniWeaving      & \underline{3.97} & \underline{3.88} & \underline{3.87} & \underline{3.91} & \underline{3.88} & \underline{3.88} & \textbf{3.67} & \underline{3.81} & \underline{3.81} & \underline{3.69} & \underline{3.60} & \underline{3.70} \\
\textbf{\ours} & \textbf{4.00} & \textbf{3.90} & \textbf{3.90} & \textbf{3.93} & \textbf{4.00} & \textbf{4.00} & \underline{3.56} & \textbf{3.85} & \textbf{4.00} & \textbf{3.86} & \textbf{3.71} & \textbf{3.86} \\
\bottomrule
\end{tabular}
\caption{Full quantitative comparison on IntelligentVBench TI2V / I2V generation subtasks.
IF: Instruction Following, CP: Condition Preserving, VQ: Visual Quality, and AVG: average score.
Best results are \textbf{bold}; second best results are \underline{underlined}.}
\label{tab:supp_intelligent_vbench}
\end{table*}

\begin{table*}[tb]
\centering
\small
\setlength{\tabcolsep}{3pt}
\begin{tabular}{l|cccccccc|c}
\toprule
 & \multicolumn{8}{c|}{\textbf{Editing Type}} & \\
\textbf{Method} & \makecell{Global\\Style} & \makecell{BG\\Change} & \makecell{Local\\Change} & \makecell{Local\\Remove} & \makecell{Local\\Add} & \makecell{Camera\\Edit} & \makecell{Creative\\Edit} & \makecell{Subtitle\\Edit} & AVG$\uparrow$ \\
\midrule
VACE             & 2.40 & 1.40 & 1.25 & 1.50 & 1.00 & 1.60 & 2.00 & 1.00 & 1.52 \\
Bernini          & \underline{4.40} & 3.02 & 4.40 & \textbf{3.33} & 3.00 & \textbf{4.80} & \underline{3.73} & \underline{2.53} & \underline{3.65} \\
Kiwi-Edit        & 4.07 & 2.87 & 3.96 & \underline{3.17} & 3.17 & 1.67 & 2.80 & \underline{2.53} & 3.03 \\
Omni-Video~2     & 3.80 & \underline{3.11} & \underline{4.42} & 1.92 & \underline{3.50} & 3.11 & 3.47 & 1.80 & 3.14 \\
OmniWeaving      & 4.07 & 2.80 & 3.96 & 2.33 & 2.83 & 1.60 & 3.13 & 1.87 & 2.82 \\
\textbf{\ours} & \textbf{4.44} & \textbf{3.40} & \textbf{4.54} & \textbf{3.33} & \textbf{3.67} & \underline{3.82} & \textbf{4.20} & \textbf{3.24} & \textbf{3.83} \\
\bottomrule
\end{tabular}
\caption{Full quantitative comparison on OpenVE-Bench.
For each editing type, each cell reports AVG across Instruction Compliance, Consistency and Detail Fidelity, and Visual Quality and Stability.
The last column reports AVG over the eight editing types.
Best results are \textbf{bold}; second best results are \underline{underlined}.}
\label{tab:supp_openve}
\end{table*}

\begin{table*}[tb]
\centering
\small
\setlength{\tabcolsep}{2pt}
\begin{tabular}{l|cccc|cccc|cccc}
\toprule
 & \multicolumn{4}{c|}{\textbf{RefVIE-Subject}} & \multicolumn{4}{c|}{\textbf{RefVIE-Background}} & \multicolumn{4}{c}{\textbf{IntelligentVBench TIV2V}} \\
\textbf{Method} & Identity & Temporal & Physical & AVG$\uparrow$ & Ref. Fid. & Matting & Visual Harm. & AVG$\uparrow$ & IF$\uparrow$ & CP$\uparrow$ & VQ$\uparrow$ & AVG$\uparrow$ \\
\midrule
VACE             & 1.00 & 1.00 & 1.00 & 1.00 & 1.40 & 1.40 & 1.40 & 1.40 & 1.89 & 2.11 & 2.78 & 2.26 \\
Bernini          & \underline{4.00} & \underline{4.00} & \underline{3.83} & \underline{3.94} & \underline{4.00} & \underline{3.60} & \textbf{3.20} & \underline{3.60} & \underline{4.11} & \underline{4.00} & \underline{4.00} & \underline{4.04} \\
Kiwi-Edit        & 3.50 & 2.67 & 2.83 & 3.00 & 2.80 & 2.40 & 2.40 & 2.53 & 3.89 & 3.78 & 3.67 & 3.78 \\
Omni-Video~2     & 3.17 & 3.17 & 3.17 & 3.17 & 2.40 & 2.20 & 2.20 & 2.27 & 3.00 & 2.33 & 3.22 & 2.85 \\
OmniWeaving      & 3.33 & 3.17 & 2.83 & 3.11 & 3.80 & 3.40 & 2.80 & 3.33 & 4.00 & 3.78 & 3.89 & 3.89 \\
\textbf{\ours} & \textbf{4.50} & \textbf{4.39} & \textbf{4.39} & \textbf{4.43} & \textbf{4.47} & \textbf{3.80} & \underline{3.07} & \textbf{3.78} & \textbf{4.30} & \textbf{4.30} & \textbf{4.19} & \textbf{4.26} \\
\bottomrule
\end{tabular}
\caption{Full quantitative comparison on RefVIE-Bench and IntelligentVBench TIV2V.
For RefVIE, AVG denotes the average over the three task-specific dimensions in each category.
RefVIE-Subject columns abbreviate Identity Consistency and Compliance (Identity), Temporal Consistency and Texture Fidelity (Temporal), and Physical Integration and Tracking (Physical); RefVIE-Background columns abbreviate Reference Fidelity and Preservation (Ref.\ Fid.), Matting Quality and Temporal Stability (Matting), and Visual Harmony and Perspective (Visual Harm.).
For TIV2V, we report IF, CP, VQ, and AVG.
Best results are in \textbf{bold}; second best results are \underline{underlined}.}
\label{tab:supp_refvie_tiv2v}
\end{table*}

% ===========================================================================
\section{Additional Qualitative Comparisons}
\label{supp:gallery}

\begin{figure*}[t]
    \centering
    \begin{subfigure}[b]{1\linewidth}
        \centering
        \includegraphics[width=\linewidth]{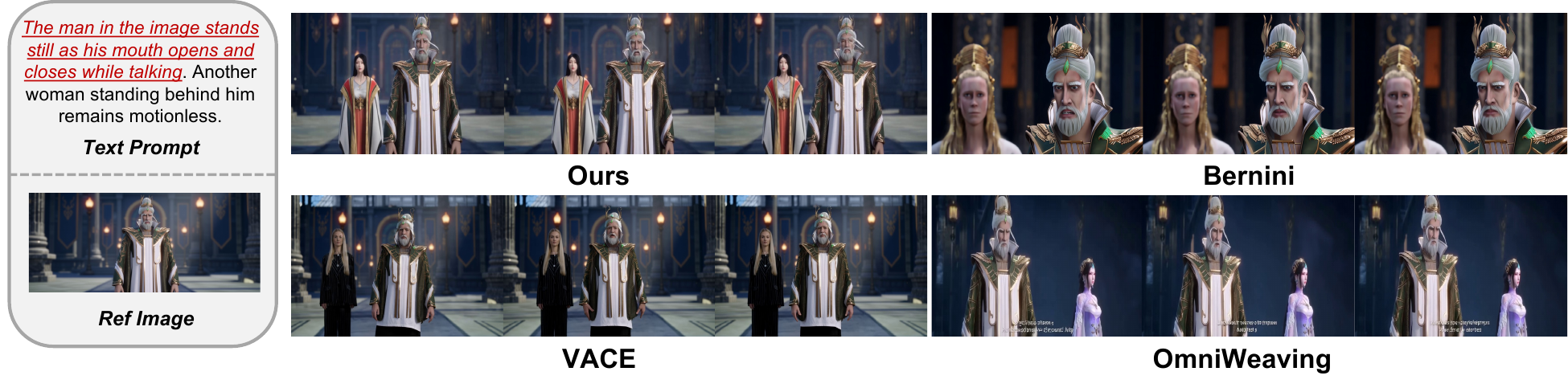}
    \end{subfigure}

    \vspace{6pt}

    \begin{subfigure}[b]{1\linewidth}
        \centering
        \includegraphics[width=\linewidth]{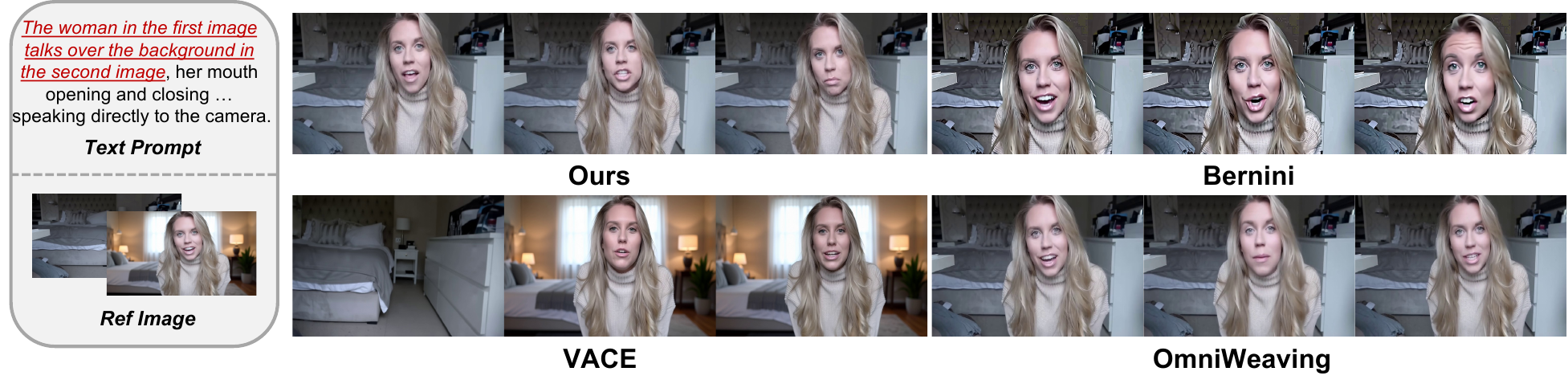}
    \end{subfigure}
    \caption{Additional qualitative comparisons on TI2V / I2V-style generation tasks from IntelligentVBench.}
    \label{fig:supp_qual_ti2v}
\end{figure*}

\begin{figure*}[t]
    \centering
    \begin{subfigure}[b]{1\linewidth}
        \centering
        \includegraphics[width=\linewidth]{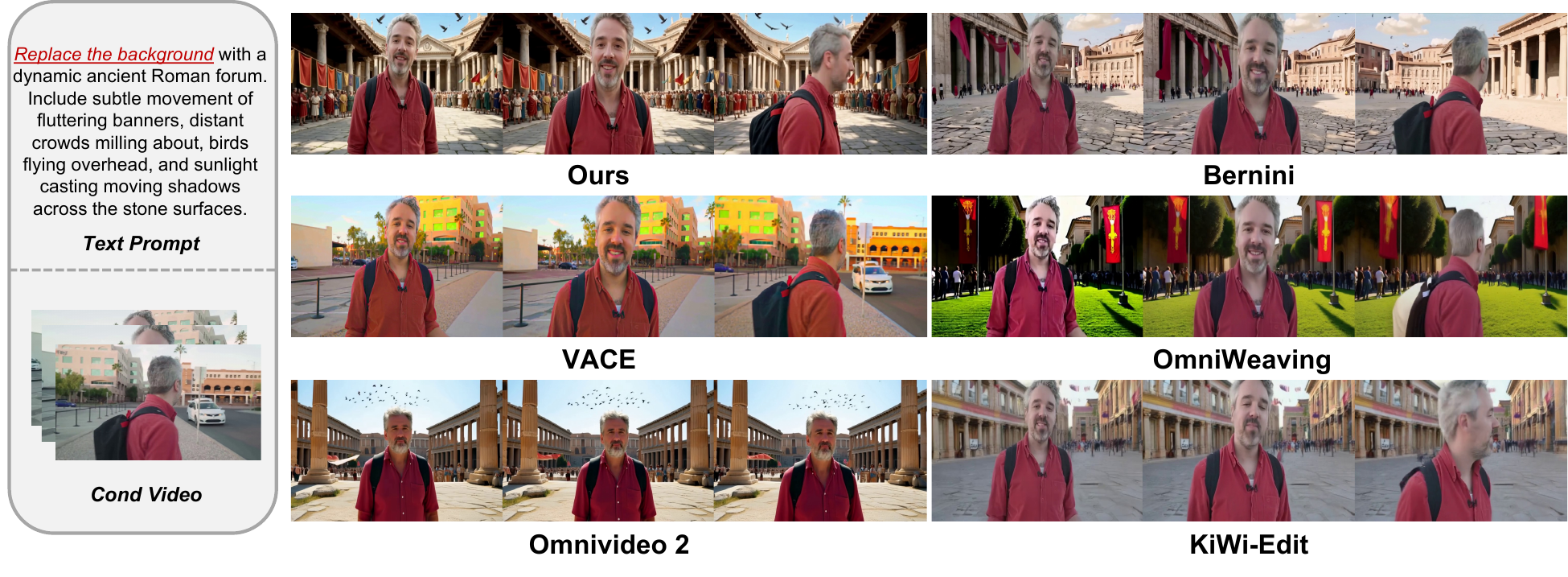}
    \end{subfigure}

    \vspace{6pt}

    \begin{subfigure}[b]{1\linewidth}
        \centering
        \includegraphics[width=\linewidth]{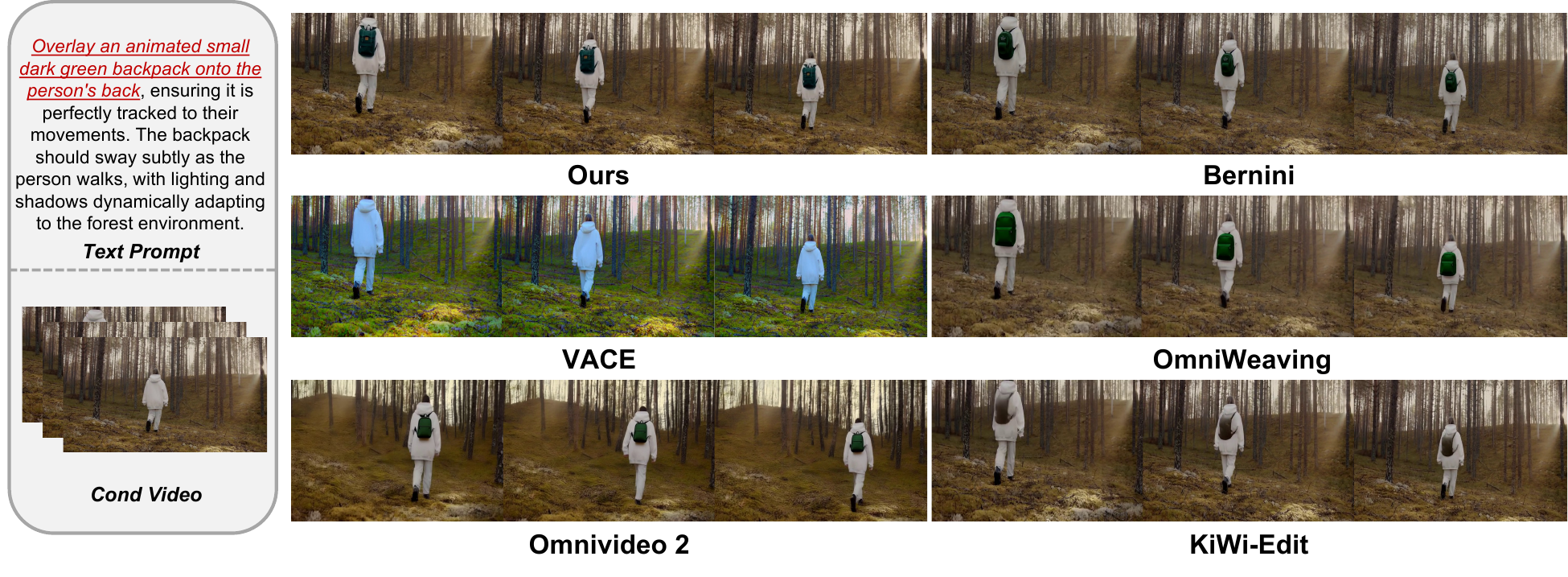}
    \end{subfigure}
    \caption{Additional qualitative comparisons on TV2V editing tasks from OpenVE-Bench.}
    \label{fig:supp_qual_tv2v}
\end{figure*}

\begin{figure*}[t]
    \centering
    \begin{subfigure}[b]{1\linewidth}
        \centering
        \includegraphics[width=\linewidth]{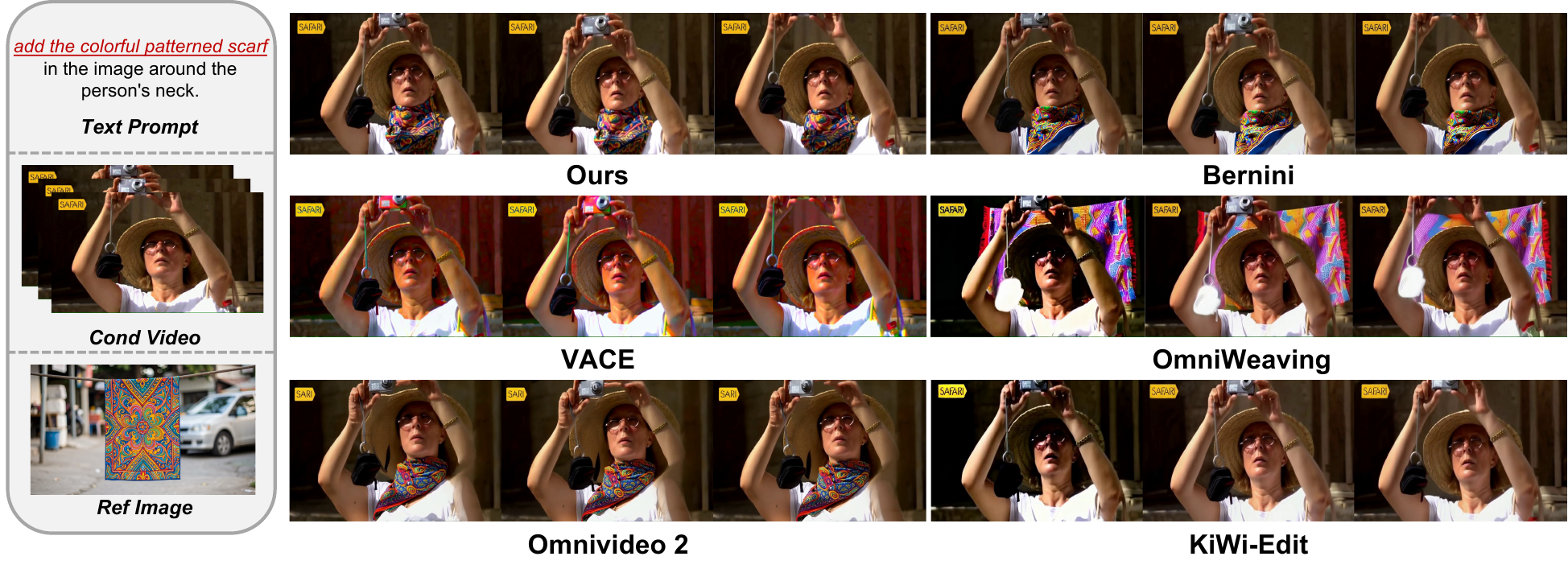}
    \end{subfigure}

    \vspace{6pt}

    \begin{subfigure}[b]{1\linewidth}
        \centering
        \includegraphics[width=\linewidth]{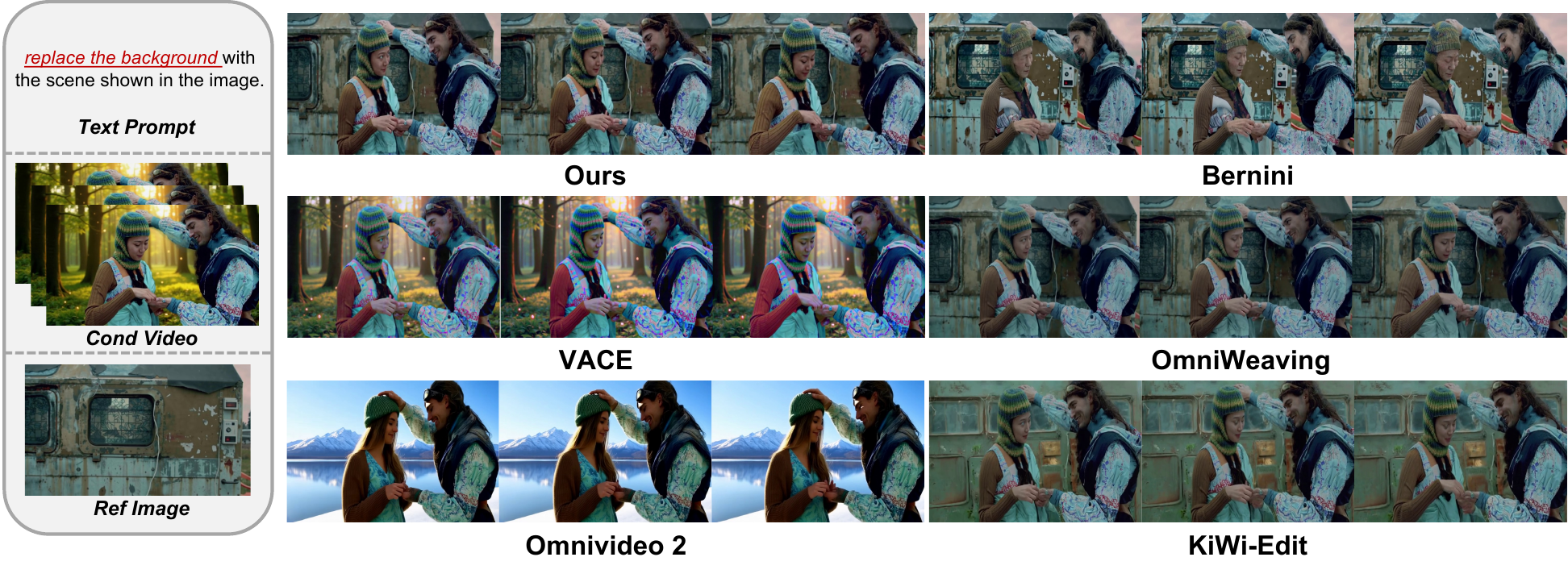}
    \end{subfigure}
    \caption{Additional qualitative comparisons on TIV2V generation/editing tasks from RefVIE-Bench and IntelligentVBench.}
    \label{fig:supp_qual_tiv2v}
\end{figure*}

\begin{figure*}[t]
    \centering
    \includegraphics[width=1\textwidth]{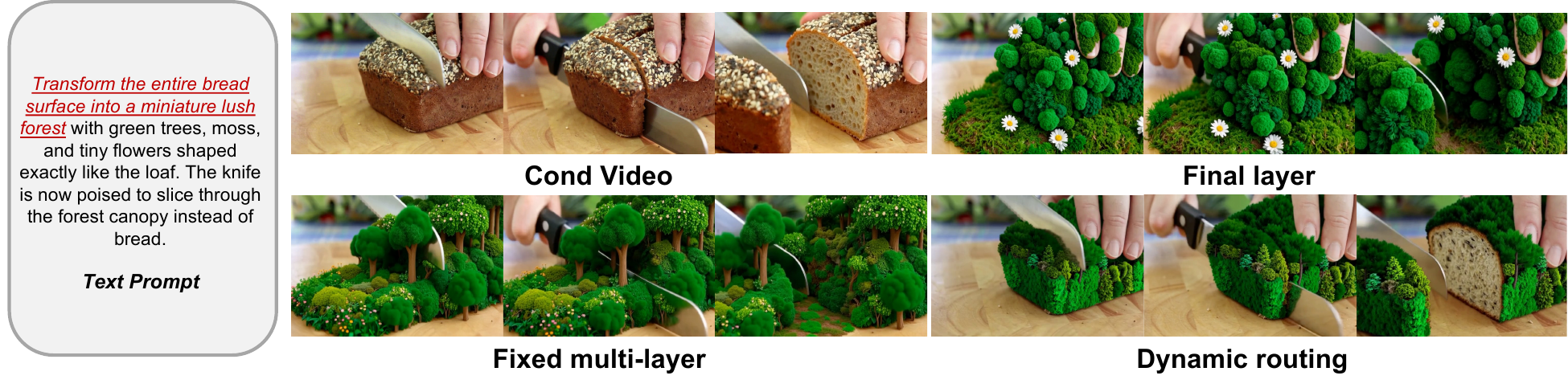}
    \caption{Additional ablation on routing strategies for a TV2V editing case.
    The case shows the prompt and conditions (left), the source video (Cond Video), and the outputs of Final layer, Fixed multi-layer, and Dynamic routing (Ours).}
    \label{fig:supp_ablation}
\end{figure*}

We provide additional qualitative cases for each task family.
The representative case for each family appears in
Figure~\ref{fig:qual_main}; the remaining cases are shown in
Figures~\ref{fig:supp_qual_ti2v}--\ref{fig:supp_qual_tiv2v} for the three task
families and in Figure~\ref{fig:supp_ablation} for the routing ablation.
Each case compares \ours with baselines using the first, middle, and last frames.

% ===========================================================================
\section{Routing Analysis}
\label{supp:routing_analysis}

Sec.~\ref{sec:ablation} attributes \ours's gains to input-conditioned layer selection.
Here we analyze the learned routing statistically and ask whether it (i) varies across tasks and (ii) varies across inputs within a task.

\begin{figure*}[t]
    \centering
    \includegraphics[width=0.99\textwidth]{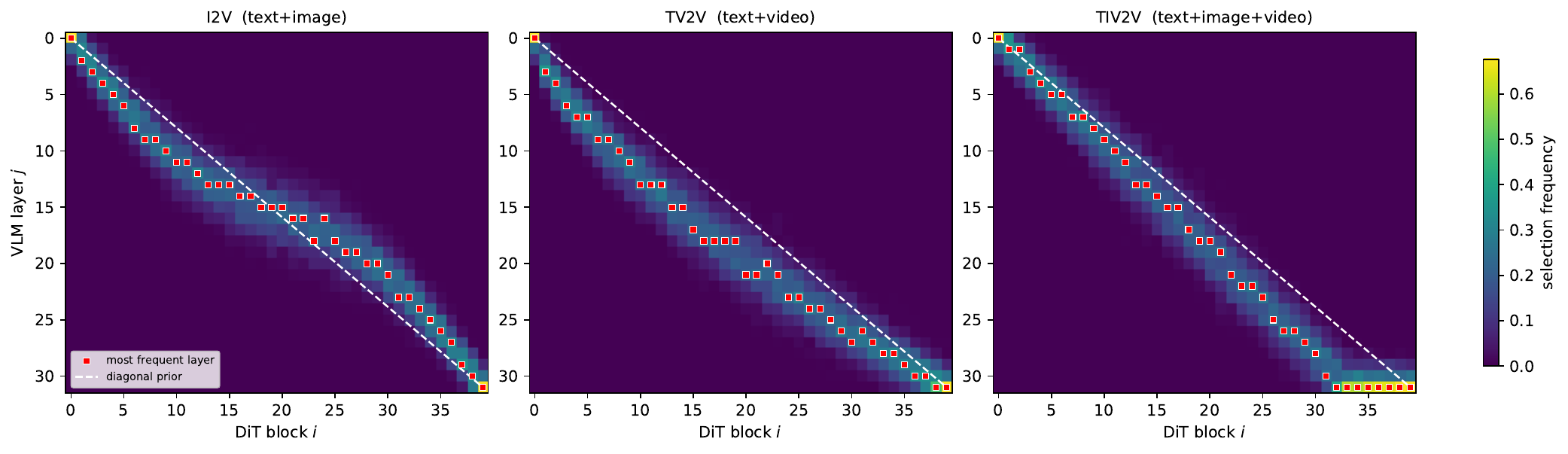}
    \caption{\textbf{Per-task average routing.}
    Selection-frequency maps over $L_D{=}40$ DiT blocks (horizontal axis) and $L_Q{=}32$ VLM layers (vertical axis), aggregated over the $N{=}225$ conditions of each task; each cell is the fraction of conditions in which a block selects a layer, and the red marker gives the most frequently selected layer for each block.
    The dashed line is the diagonal Gaussian prior used at initialization (Sec.~\ref{supp:gaussian_prior}).
    All three tasks keep a near-diagonal ordering but bend away from the prior in different ways; the ordering is nonetheless learned rather than imposed, as removing the prior from the final logits changes only 0.8--1.1\% of the decisions.}
    \label{fig:supp_routing_task}
\end{figure*}

\begin{figure*}[t]
    \centering
    \includegraphics[width=0.86\textwidth]{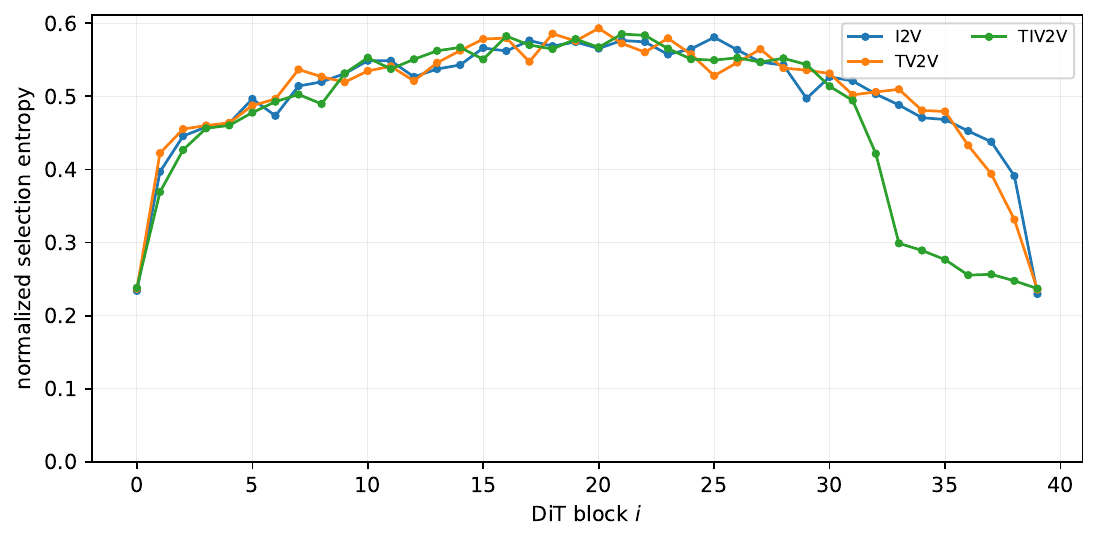}
    \caption{\textbf{Within-task routing variability.}
    Normalized entropy of the selected VLM layer across the conditions of one task, per DiT block.
    Zero entropy would indicate a fixed block-to-layer mapping.
    Entropy peaks at mid-depth blocks and drops at both ends; for TIV2V it collapses after block~32, where the modal choice saturates at the last VLM layer.}
    \label{fig:supp_routing_entropy}
\end{figure*}

\paragraph{Setup.}
We collect routing decisions for $N{=}225$ conditions per task.
The conditions combine samples from the benchmark test sets with additional conditions assembled from the training corpus, as the test sets alone are too small for the statistics below.
Of the 225 conditions per task, 200 share their underlying content across all three tasks and differ only in the condition composition (text${+}$image for TI2V, text${+}$video for TV2V, text${+}$image${+}$video for TIV2V); this paired subset separates the effect of the condition composition from that of the content.
Routing depends only on the condition, not on the noise or the denoising step, so each decision is obtained with a single condition-only forward pass and no video generation; by construction, the route is independent of the generation seed.
For each task we compute a selection-frequency map $\mathbf{M} \in [0,1]^{L_D \times L_Q}$, where $M_{ij}$ is the fraction of conditions in which DiT block $i$ selects VLM layer $j$ (i.e., $\ell_i = j$), with $L_D{=}40$ and $L_Q{=}32$.
Blocks and layers are numbered from zero in Figures~\ref{fig:supp_routing_task} and~\ref{fig:supp_routing_entropy}.

\paragraph{Routing varies across tasks.}
All three tasks keep a shallow-to-deep ordering but distribute the VLM layers differently (Figure~\ref{fig:supp_routing_task}).
The mean per-block Jensen--Shannon divergence between task maps is 0.27--0.43 bits, an order of magnitude above both the split-half floor within a task (0.025--0.028) and a task-label permutation null (0.016--0.018; $z > 149$ for every pair).
On the paired subset, where only the condition composition changes, the selected layer still shifts by 2.1--3.1 layers, and matching the content at random instead of by identity alters this gap by only 6--11\%, so the composition rather than the content drives the difference.

\paragraph{Routing is input-dependent.}
The mean normalized selection entropy is 0.50 for TI2V, 0.50 for TV2V, and 0.47 for TIV2V, i.e.\ five to six effective candidate layers per block, and every one of the $L_Q{=}32$ layers is selected by some block: neither the per-input choice nor the aggregate layer usage collapses.
Entropy is lowest at the first and last blocks ($\approx$0.23) and highest at mid-depth ($\approx$0.58); for TIV2V it falls from 0.52 to 0.29 after block~32, where the modal choice saturates and little room for variation remains (Figure~\ref{fig:supp_routing_entropy}).

% ===========================================================================
\section{Statement on the Use of LLMs}
\label{supp:llm_statement}

For transparency, we disclose the use of large language models (LLMs) as assistive tools in preparing this paper.
We used OpenAI GPT~5.6 for language editing and refinement.
The scope of its use was strictly confined to improving the manuscript's readability, including tasks such as grammar correction, sentence restructuring for clarity, and style enhancements.
We state that the LLM was not involved in the ideation of the research, the formulation of the methodology, the generation or analysis of experimental results, or the drawing of scientific conclusions.
Separately, Gemini-2.5-Pro serves as the automatic evaluator in our benchmark protocol; this is a component of the evaluation methodology, is fully described in Sec.~\ref{sec:main_results} and Sec.~\ref{supp:full_tables}, and is distinct from the writing assistance disclosed here.